\documentclass[twoside,twocolumn,9pt]{article}
\usepackage{extsizes}
\usepackage[super,sort&compress,comma]{natbib} 
\usepackage[version=3]{mhchem}
\usepackage[left=1.5cm, right=1.5cm, top=1.785cm, bottom=2.0cm]{geometry}
\usepackage{balance}
\usepackage{mathptmx}
\usepackage{sectsty}
\usepackage{graphicx} 
\usepackage{lastpage}
\usepackage[format=plain,justification=justified,singlelinecheck=false,font={stretch=1.125,small,sf},labelfont=bf,labelsep=space]{caption}
\usepackage{float}
\usepackage{fancyhdr}
\usepackage{fnpos}
\usepackage[english]{babel}
\addto{\captionsenglish}{%
  
}
\usepackage{array}
\usepackage{droidsans}
\usepackage{charter}
\usepackage[T1]{fontenc}
\usepackage[usenames,dvipsnames]{xcolor}
\usepackage{setspace}
\usepackage[compact]{titlesec}
\usepackage{hyperref}
\usepackage{microtype}
\usepackage{amsmath}
\usepackage{amsfonts}
\usepackage{amssymb}
\usepackage{threeparttable}
\usepackage{soul}
\usepackage{mdframed}
\usepackage{fvextra}
\usepackage{booktabs}
\usepackage{longtable}
\usepackage{multirow}
\usepackage{xspace}
\usepackage{diagbox}
\usepackage{makecell}
\usepackage{fancyvrb}

\definecolor{cream}{RGB}{222,217,201}
\definecolor{darkblue}{rgb}{0, 0, 0.5}
\hypersetup{colorlinks=true, citecolor=darkblue, linkcolor=darkblue, urlcolor=darkblue}
\newcommand{\framework}{\mbox{$\mathop{\textsc{MMORF}}\limits$}\xspace}

\newcommand{\masil}{\mbox{$\mathop{\textsc{MASIL}}\limits$}\xspace}

\newcommand{\rfas}{\mbox{$\mathop{\textsc{RFAS}}\limits$}\xspace}

\newcommand{\coordinator}{\mbox{$\mathop{\textsc{Coordinator}}\limits$}\xspace}

\newcommand{\rules}{\mbox{$\mathop{\textsc{Regulator}}\limits$}\xspace}

\newcommand{\steering}{\mbox{$\mathop{\textsc{Navigator}}\limits$}\xspace}

\newcommand{\verifier}{\mbox{$\mathop{\textsc{Verifier}}\limits$}\xspace}

\newcommand{\tools}{\mbox{$\mathop{\textsc{Tooling}}\limits$}\xspace}

\newcommand{\meea}{\mbox{$\mathop{\textsc{MEEA$^{\!*}$}}\limits$}\xspace}

\newcommand{\tango}{\mbox{$\mathop{\textsc{Tango$^{*}$}}\limits$}\xspace}
\newcommand{\desp}{\mbox{$\mathop{\textsc{DESP}}\limits$}\xspace}
\newcommand{\larc}{\mbox{$\mathop{\textsc{LARC}}\limits$}\xspace}

\newcommand{\valuefn}{\mbox{$\mathop{V}\limits$}\xspace}

\newcommand{\turnlimit}{\mbox{$\mathop{T}\limits$}\xspace}

\newcommand{\presence}{\mbox{$\mathop{\texttt{PR}}\limits$}\xspace}

\newcommand{\validity}{\mbox{$\mathop{\texttt{VR}}\limits$}\xspace}

\newcommand{\success}{\mbox{$\mathop{\texttt{SR}}\limits$}\xspace}

\newcommand{\ps}{\mbox{$\mathop{\texttt{P-S}}\limits$}\xspace}

\newcommand{\SP}{\mbox{$\mathop{\texttt{S}/\texttt{P}}\limits$}\xspace}

\newcommand{\carcinogenicity}{\mbox{$\mathop{\texttt{Carc}}\limits$}\xspace}

\newcommand{\pyrophoricity}{\mbox{$\mathop{\texttt{Pyro}}\limits$}\xspace}

\newcommand{\ghs}{\mbox{$\mathop{\texttt{GHS}}\limits$}\xspace}

\newcommand{\totalcost}{\mbox{$\mathop{\texttt{SMP}}\limits$}\xspace}

\newcommand{\routelength}{\mbox{$\mathop{\texttt{RL}}\limits$}\xspace}

\newcommand{\qwen}{\mbox{$\mathop{\textsc{Qwen3 30B Instruct}}\limits$}\xspace}

\newcommand{\qwenEightyB}{\mbox{$\mathop{\textsc{Qwen3 Next 80B Instruct}}\limits$}\xspace}

\newcommand{\mistral}{\mbox{$\mathop{\textsc{Mistral Nemo 12B}}\limits$}\xspace}

\newcommand{\claude}{\mbox{$\mathop{\textsc{Claude 4.5 Sonnet}}\limits$}\xspace}

\newcommand{\gpt}{\mbox{$\mathop{\textsc{GPT 5.1}}\limits$}\xspace}

\newcommand{\kimi}{\mbox{$\mathop{\textsc{Kimi K2.5}}\limits$}\xspace}

\newcommand{\deepseek}{\mbox{$\mathop{\textsc{DeepSeek R1}}\limits$}\xspace}

\newcommand{\pareto}{\mbox{$\mathop{\textsc{Pareto}_{10}}\limits$}\xspace}

\newcommand{\feasibility}{\mbox{$\mathop{\texttt{Feas}}\limits$}\xspace}

\newcommand{\static}{\mbox{$\mathop{\textsc{StaticReg}}\limits$}\xspace}

\newcommand{\gmo}{\mbox{$\mathop{\texttt{SCMO}\text{-retro}}\limits$}\xspace}

\newcommand{\cmo}{\mbox{$\mathop{\texttt{HCMO}\text{-retro}}\limits$}\xspace}

\newmdenv[
  backgroundcolor=gray!5,
  linecolor=gray!50,
  skipabove=6pt,
  skipbelow=6pt,
  innertopmargin=6pt,
  innerbottommargin=6pt,
  innerleftmargin=6pt,
  innerrightmargin=6pt,
  font=\small
]{promptbox}

\newcommand{\vardef}[1]{\texttt{\{#1\}}}

\begin{document}

\pagestyle{fancy}
\thispagestyle{plain}
\fancypagestyle{plain}{
\renewcommand{\headrulewidth}{0pt}
}

\makeFNbottom
\makeatletter
\renewcommand\LARGE{\@setfontsize\LARGE{15pt}{17}}
\renewcommand\Large{\@setfontsize\Large{12pt}{14}}
\renewcommand\large{\@setfontsize\large{10pt}{12}}
\renewcommand\footnotesize{\@setfontsize\footnotesize{7pt}{10}}
\makeatother

\renewcommand{\thefootnote}{\fnsymbol{footnote}}
\renewcommand\footnoterule{\vspace*{1pt}%
\color{cream}\hrule width 3.5in height 0.4pt \color{black}\vspace*{5pt}} 
\setcounter{secnumdepth}{5}

\makeatletter 
\renewcommand\@biblabel[1]{#1}            
\renewcommand\@makefntext[1]%
{\noindent\makebox[0pt][r]{\@thefnmark\,}#1}
\makeatother 
\renewcommand{\figurename}{\small{Fig.}~}
\sectionfont{\sffamily\Large}
\subsectionfont{\normalsize}
\subsubsectionfont{\bf}
\setstretch{1.125} 
\setlength{\skip\footins}{0.8cm}
\setlength{\footnotesep}{0.25cm}
\setlength{\jot}{10pt}
\titlespacing*{\section}{0pt}{4pt}{4pt}
\titlespacing*{\subsection}{0pt}{15pt}{1pt}

\fancyfoot{}
\fancyfoot[LO,RE]{\vspace{-7.1pt}Preprint. Under review.}
\fancyfoot[CO]{\vspace{-7.1pt}\hspace{13.2cm}}
\fancyfoot[CE]{\vspace{-7.2pt}\hspace{-14.2cm}}
\fancyfoot[RO]{\footnotesize{\sffamily{1--\pageref{LastPage} ~\textbar  \thepage}}}
\fancyfoot[LE]{\footnotesize{\sffamily{\thepage~\textbar1--\pageref{LastPage}}}}
\fancyhead{}
\renewcommand{\headrulewidth}{0pt} 
\renewcommand{\footrulewidth}{0pt}
\setlength{\arrayrulewidth}{1pt}
\setlength{\columnsep}{6.5mm}
\setlength\bibsep{1pt}

\makeatletter 
\newlength{\figrulesep} 
\setlength{\figrulesep}{0.5\textfloatsep} 

\newcommand{\topfigrule}{\vspace*{-1pt}%
\noindent{\color{cream}\rule[-\figrulesep]{\columnwidth}{1.5pt}} }

\newcommand{\botfigrule}{\vspace*{-2pt}%
\noindent{\color{cream}\rule[\figrulesep]{\columnwidth}{1.5pt}} }

\newcommand{\dblfigrule}{\vspace*{-1pt}%
\noindent{\color{cream}\rule[-\figrulesep]{\textwidth}{1.5pt}} }

\makeatother

\twocolumn[
  \begin{@twocolumnfalse}
\vspace{1em}
\sffamily
\begin{tabular}{p{18cm} }

\noindent\LARGE{\textbf{\framework: A Multi-agent Framework for Designing \mbox{Multi-objective} \mbox{Retrosynthesis} Planning Systems}} \\
 \vspace{0.3cm} \\

 \noindent\large{Frazier N. Baker$^{a}$, Trieu Nguyen$^{a}$, Reza Averly$^{a}$, Botao Yu$^{a}$, Daniel Adu-ampratwum$^{b}$, Huan Sun$^{a}$, and Xia Ning$^{a,b,c,d}$} \\
\vspace{0.3cm} \\
\noindent\normalsize{
Multi-objective retrosynthesis planning is
a critical chemistry task requiring dynamic balancing of quality, safety, and cost objectives.
Language model-based
multi-agent systems (MAS) offer a promising approach for this task:
leveraging interactions of
specialized agents to incorporate multiple objectives into retrosynthesis planning.
We present \framework, a framework for constructing MAS for multi-objective retrosynthesis planning.
\framework features modular agentic components, 
which can be flexibly combined and configured into different systems,
enabling principled evaluation and comparison of different system designs.
Using \framework, we construct two representative MAS: \masil and \rfas.
On a newly curated benchmark consisting of 218 multi-objective retrosynthesis planning tasks,
\masil achieves strong safety and cost metrics on soft-constraint tasks, frequently Pareto-dominating baseline routes,
while \rfas achieves a 48.6\% success rate on hard-constraint tasks, outperforming state-of-the-art baselines.
Together, these results show the effectiveness of \framework as a foundational framework for exploring MAS for multi-objective retrosynthesis planning.
Code and data are available at \url{https://github.com/ninglab/MMORF}.
} \\

\end{tabular}

 \end{@twocolumnfalse} \vspace{0.6cm}

  ]

\renewcommand*\rmdefault{bch}\normalfont\upshape
\rmfamily
\section*{}
\vspace{-1cm}


\footnotetext{\textit{$^{a}$~Department of Computer Science and Engineering, The Ohio State University, 395 Dreese Laboratories, 2015 Neil Avenue, Columbus, Ohio, United States of America.}}
\footnotetext{\textit{$^{b}$~Division of Medicinal Chemistry and Pharmacognosy, The Ohio State University, 217 Lloyd M. Parks Hall, 500 West 12th Avenue, Columbus, Ohio, United States of America.}}
\footnotetext{\textit{$^{c}$~Department of Biomedical Informatics, The Ohio State University, 250 Lincoln Tower, 1800 Cannon Drive, Columbus, Ohio, United States of America.}}
\footnotetext{\textit{$^{d}$~Translational Data Analytics Institute, The Ohio State University, 175 Pomerene Hall, 1760 Neil Avenue, Columbus, Ohio, United States of America.}}

\footnotetext{\dag~Supplementary Information (SI) available: [details of any supplementary information available should be included here]. See DOI: 00.0000/00000000.}



\section{Introduction}
\label{sec:intro}

Retrosynthesis planning is a critical task in chemistry~\citep{jiang_artificial_2023,butters_route_2011},
requiring the identification of a synthetic route for a given target molecule (product) from commercially available starting materials.
This task is fundamentally challenging, as it requires searching the vast space of possible chemical reactions.
Multi-objective retrosynthesis further increases the difficulty of the task by considering multiple objectives,
such as minimizing route length, avoiding carcinogenic substances, or limiting the cost of starting materials.
Dynamically balancing such objectives requires strong, principled reasoning grounded in chemistry,
as every decision made during retrosynthesis planning can impact multiple objectives simultaneously.
In this setting, large language model (LLM)-based multi-agent systems (MAS)
offer a promising approach:
leveraging interactions among specialized agents to incorporate multiple objectives into retrosynthesis planning.


Artificial intelligence (AI) for multi-objective retrosynthesis planning is an emerging area of study.
Traditionally, AI methods for retrosynthesis planning focus on a single objective:
identifying a valid synthetic route.
Recent efforts~\citep{larc, desp, tango} have explored hard-constraint multi-objective retrosynthesis planning (\cmo),
where, in addition to identifying a valid route, explicit, user-specified requirements must be satisfied as additional objectives.
Specifically, existing work focuses on {\cmo} with a single constraint.
However, such formulations do not capture more realistic settings, 
such as \cmo with multiple constraints, or
soft-constraint multi-objective retrosynthesis planning (\gmo)---where
many objectives must be dynamically balanced throughout planning. 
MAS are well-suited to these challenging tasks, as they can employ specialized agents that reason over multiple objectives and interact to shape retrosynthesis planning through distinct strategies.
Furthermore, MAS are inherently modular, facilitating the exploration of different system design strategies to fit a variety of practical settings.

\begin{figure*}

    \centering
\includegraphics[width=0.95\linewidth]{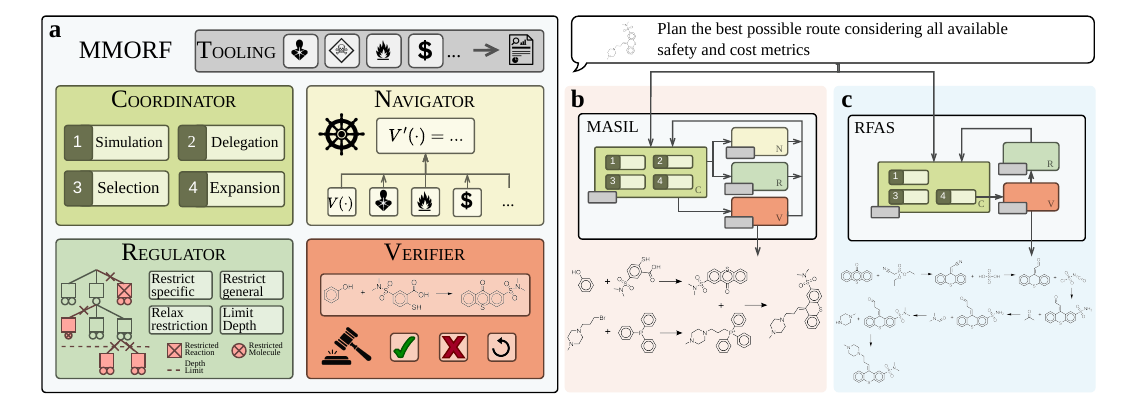} 
    \caption{Overview of \textbf{a,} \framework, and two MAS built using it: \textbf{b,} \masil, and \textbf{c,} \rfas.}
    \label{fig:overview}
\end{figure*}

Here, we present \framework (\textbf{M}ulti-agent \textbf{M}ulti-\textbf{O}bjective \textbf{R}etrosynthesis \textbf{F}ramework), a framework for constructing MAS for multi-objective retrosynthesis planning.
Figure~\ref{fig:overview}\textbf{a} shows an overview of \framework.
\framework features four modular agentic components,
which can be flexibly combined and configured into different systems:
\textbf{(1)} \coordinator coordinates the overall retrosynthesis planning process;
\textbf{(2)} \steering
calibrates multiple objective-specific signals into a unified guiding function;
\textbf{(3)} \rules controls the boundaries of the retrosynthesis planning space;
and
\textbf{(4)} \verifier judges whether a route can be returned or if further planning is needed.
These components can be flexibly combined and configured into many different systems;
we design two distinct, representative MAS for evaluation and comparison to illustrate the impacts of different design choices under \framework:
\masil (Figure~\ref{fig:overview}\textbf{b}),
which tightly integrates multi-agent reasoning into the entire retrosynthesis planning process;
and \rfas (Figure~\ref{fig:overview}\textbf{c}),
which selectively incorporates multi-agent guidance during key decision points in retrosynthesis planning.
\masil and \rfas are, to the best of our knowledge, the first MAS for multi-objective retrosynthesis,
establishing \framework as a foundational framework for this new paradigm in multi-objective retrosynthesis planning.

We evaluate \masil and \rfas on a newly curated benchmark consisting of 218 multi-objective retrosynthesis planning tasks.
This dataset is the first of its kind,
comprising both \cmo and \gmo tasks.
\masil achieves strong results on safety and cost metrics on \gmo tasks, frequently Pareto-dominating baseline routes,
whereas \rfas achieves a 48.6\% success rate on \cmo tasks, outperforming state-of-the-art \cmo baselines.
Further analysis of these results illustrates how specific \framework components
and design choices affect multi-objective retrosynthesis planning performance.
A reaction feasibility evaluation shows that reactions in routes produced by 
\masil and \rfas are significantly more likely to be feasible than reactions from LLM baseline routes,
highlighting the importance of the principled chemistry underlying \framework components.
We conduct a case study on \masil showing how all the agentic components of \framework can work together to produce routes with strong safety and cost profiles. 
Together, these results demonstrate the effectiveness of \framework as a foundational framework for exploring MAS for multi-objective retrosynthesis planning.
The code and data for \framework are available at \url{https://github.com/ninglab/MMORF}.

\section{Related Work}
\label{sec:related_work}

Artificial intelligence (AI) for retrosynthesis planning~\citep{meea,retrostar,sirp,genheden_aizynthfinder_2020,segler_planning_2018}
traditionally focuses on a single objective:
finding a synthetic route for the target molecule (product).
Recent efforts~\citep{larc, desp, tango,bran_chemical_2025, xuan-vu_synthelite_2025} have explored \cmo tasks
with a single, user-specified constraint as the additional objective.
\larc~\citep{larc} constrains retrosynthesis planning using an LLM-based Agent-as-a-Judge, steering planning away from reactions with hazardous molecules.
\desp~\citep{desp} and \tango~\citep{tango} constrain retrosynthesis planning using molecule similarity to guide planning towards desired starting materials.
Other methods~\citep{bran_chemical_2025, xuan-vu_synthelite_2025} constrain retrosynthesis planning based on reaction mechanisms,
using LLMs to guide retrosynthesis planning toward reactions that align with chemists' preferences.
However, these efforts  
are formulated around one single, specific constraint, 
whereas in practice,
synthetic chemists must dynamically balance many objectives simultaneously,
often lacking strict, well-defined constraints~\citep{butters_route_2011}.

For \gmo tasks,
\textsc{MO-MCTS}~\citep{lai_multi_2025} plans routes using a Pareto-based multi-objective version of Monte Carlo Tree Search (MCTS) with objectives focused on basic route quality (e.g., short route length) and synthetic strategy (e.g., matching a reference route).
However, \textsc{MO-MCTS} fixes its Pareto-based MCTS strategy at the outset,
limiting its ability to dynamically adapt or reason over the relative importance of different objectives.
Across all methods discussed, each proposes only a single, fixed mechanism to incorporate multiple objectives into retrosynthesis planning,
limiting their flexibility and generalizability to optimally
handle many, complicated objectives simultaneously.
These limitations motivate a shift towards a flexible framework for designing intelligent, adaptive multi-objective retrosynthesis planning methods.

\section{\framework Framework}
\label{sec:framework}


\framework is a framework for constructing MAS for multi-objective retrosynthesis planning.
Multi-agent systems built using \framework can effectively plan synthetic routes 
while simultaneously satisfying multiple explicit constraints or intelligently balancing implicit objectives,
such as cost (e.g., route length,
starting material price) or safety (e.g., globally harmonized system (GHS) hazard statements, pyrophoricity, carcinogenicity). 
\framework features four modular and configurable agentic components:
\textbf{(1)} \coordinator,
which directs the overall retrosynthesis planning process, optionally delegating tasks to other agents;
\textbf{(2)} \steering, which 
dynamically calibrates multiple objective-specific signals into a unified function to guide retrosynthesis planning,
\textbf{(3)} \rules, which controls the boundaries of the retrosynthesis planning space; and
\textbf{(4)} \verifier, which acts as a route-level judge,
deciding whether to return a completed route or continue planning.
Each component makes agentic decisions grounded by \tools, a non-agentic component of \framework comprised of chemistry-specific software.
The modularity of \framework allows MAS to use different combinations of its components,
omitting or including components depending on each system's specific design.
Through flexible composition and configuration of its components,
\framework enables principled evaluation and comparison of distinct system designs.
%

\subsection{\coordinator}
\label{sec:framework:coordinator}

\coordinator orchestrates the entire retrosynthesis planning process,
iteratively searching the vast chemical space and delegating tasks to other components.
\coordinator follows a four-step process for retrosynthesis planning:
\textbf{(1)} simulation, where \coordinator uses a value-function-based policy to 
review the retrosynthesis planning space and identify partial routes as candidates for further planning;
\textbf{(2)} delegation, where \coordinator optionally invokes other agentic components (e.g., \steering, \rules) with instructions based on the simulation candidates;
\textbf{(3)} selection, where \coordinator decides which candidate to pursue for further planning using either a value-function-based or LLM-based agentic policy;
and \textbf{(4)} expansion, where \coordinator predicts new reactions to extend the selected candidate route.
This process repeats for $I_{\max}$=500 iterations or until a route is found, at which point the \coordinator can consult \verifier to determine whether to continue.
Thus, \coordinator is the core of \framework ---its
decision-making is key to navigating the vast retrosynthesis planning space and finding high-quality synthetic routes.

The simulation-selection-expansion workflow used by \coordinator is well-established in retrosynthesis planning literature~\citep{meea, larc}.
Traditionally, simulation and selection in these workflows seek to maximize a pre-trained value function ($\valuefn$), designed to efficiently find a short, plausible retrosynthesis route.
\coordinator builds upon the strengths of this workflow,
leveraging a hybrid of LLM-based and value-function-based decision making.
By adding a delegation step, \coordinator can use agentic reasoning to dynamically adapt $\valuefn$ via \steering and the bounds of its retrosynthesis planning space via \rules,
fundamentally changing future simulations, selections, and expansions.
Furthermore, \coordinator can directly use LLM-based reasoning during selection rather than relying solely on the value function.
The balance between LLM-based and value-function-based decision-making can be configured differently for agentic systems constructed using \framework.
For example, delegation can be
disabled for the first $N$ iterations of the workflow, or disabled entirely.
Additionally, selection can be configured to always maximize the current value function 
$\valuefn$ or always use LLM-based agentic reasoning.
\coordinator can be configured to return the first route found, or consult \verifier for approval.
This flexibility underscores the power of \framework as a framework for studying many different agentic systems for multi-objective retrosynthesis planning.

\subsection{\steering}
\label{sec:framework:steering}

\steering is an LLM-based agent responsible for calibrating multiple objective-specific signals into a unified function to guide the expansion of the retrosynthesis planning space.
\steering takes a value-function-based approach to guide retrosynthesis planning, 
strategically incorporating tool-based feedback to guide future iterations of retrosynthesis planning.
Given a candidate set of partial routes from the current retrosynthesis planning space, detailed information on the routes based on \tools (Section~\ref{sec:framework:tools}),
and the current value function $\valuefn$ guiding the search,
\steering generates a new value function $\valuefn'$ to guide future selection and expansion.
\steering defines $\valuefn'$ by combining \tools-based value function terms using basic arithmetic functions: addition, subtraction, multiplication, and division.
This provides substantial flexibility in guiding the retrosynthesis search, enabling fine-grained weighting and even nonlinear combinations of terms.
Furthermore, 
\steering can use multi-turn reasoning to observe the impact of $\valuefn'$ on the candidates
and make further adjustments until it is satisfied with $\valuefn'$ or it reaches a pre-defined turn limit $\turnlimit=3$.
This allows \steering to dynamically adapt its value function generation to the specific retrosynthesis planning task through limited trial-and-error.

\subsection{\rules}
\label{sec:framework:rules}

\rules is an LLM-based agent that defines and manages regulations that restrict the boundaries of the retrosynthesis planning space.
Given a candidate set of partial or complete routes, detailed information on the routes based on \tools, and the current regulations bounding the retrosynthesis planning space,
\rules can take the following actions to strategically shape the retrosynthesis planning space:
\textbf{(1)} restrict a specific molecule or reaction 
\textbf{(2)} restrict a general pattern of reactions or molecules 
\textbf{(3)} limit the depth of the route,
or \textbf{(4)} remove or relax previous restrictions and limits.
Through {\rules}, {\framework} systems are able to directly apply hard constraints to the retrosynthesis 
planning space---dynamically eliminating undesirable directions from consideration.
This differs from the approaches taken previously~\citep{desp,larc,tango}, which rely solely on value-function-driven soft constraints or agentic guidance.
Similar to \steering,
\rules can use multi-turn reasoning to refine its regulations after observing the impact of its actions on the candidate set.
\rules's regulations can be efficiently enforced, imposing minimal overhead on the retrosynthesis planning process.

\subsection{\verifier}
\label{sec:framework:verifier}

\verifier is an LLM-based, route-level Agent-as-a-Judge that decides whether synthetic routes are of sufficient quality to terminate planning.
This provides a key feature lacking in other retrosynthesis frameworks~\citep{larc, desp, tango, meea}.
Given a complete route from the target to purchasable starting materials, \verifier must judge whether to \textbf{(1)} approve the route, returning it to the user; \textbf{(2)} reject the route, summarizing the reasons for rejection as feedback to the agentic system;
or \textbf{(3)} revisit a previously rejected route, approving it in light of new information.
\verifier makes its judgments using the LLM's intrinsic reasoning capabilities within the context of user-defined objectives, grounded by \tools.
\verifier retains a record of past judgments within the same retrosynthesis planning task, allowing it to compare routes to previously rejected routes and calibrate its decision making based on observed trends.
This also enables \verifier to retroactively approve previously rejected routes at later stages, allowing it to dynamically adapt its strategy to trends in the route planning results.

\subsection{\tools}
\label{sec:framework:tools}

\tools grounds
the agentic decision-making of every component in \framework.
Extensible by design, \tools includes computational methods for carcinogenicity prediction, 
pyrophoricity prediction, GHS statement retrieval, molecule structural similarity measurement, starting material cost estimation, and route length calculation.
\tools provides a comprehensive, multi-objective report for any route presented to \coordinator, \steering, \rules, or \verifier.
This report contains information on key objectives for multi-objective retrosynthesis planning, focusing on safety and cost metrics. 
Additionally, \tools also supports \steering,
allowing the direct incorporation cheminformatics computations into value function guidance for retrosynthesis planning.
More details about tool implementation are provided in the Supplementary Information (SI)\dag.

\section{\framework Systems}
\label{sec:methods}

Using \framework, myriad MAS can be constructed for multi-objective retrosynthesis planning.
Here, we present two systems using \framework: \masil and \rfas.
\masil tightly integrates multi-agent reasoning and guidance into the retrosynthesis planning process.
Its \coordinator operates in concert with \steering, \rules, and \verifier to ensure continuous multi-objective guidance.
\rfas incorporates multi-agent reasoning and guidance selectively during multi-objective retrosynthesis planning.
Its \coordinator engages \verifier and \rules at key decision points.
These serve as two representative examples of systems possible under \framework, each with its own distinct strengths. 
We instantiate \masil and \rfas using base models of different sizes: \mistral~\citep{mistral_ai_mistral_2024}, \qwen~\citep{qwen80b}, and \claude~\citep{claude4_5}.
In each instantiation, the same base model is used for all \framework components.
Please note that \framework is not dependent on any particular LLM size or family, and can leverage new LLMs as they emerge.
Base model details are discussed in the SI\dag. 

\subsection{\masil}
\label{sec:methods:masil}

\masil is a \textbf{M}ulti-\textbf{A}gent \textbf{S}ystem that integrates \framework components \textbf{I}nto each iteration of retrosynthesis planning \textbf{L}oop,
enabling continuous multi-objective guidance.
\masil leverages
all four agentic components of \framework: \coordinator, \steering, \rules, and \verifier.
In \masil, \coordinator is configured with an option to delegate to \steering or \rules during the delegate step.
This provides \masil with the ability to reshape both the priorities (via \steering) and boundaries (via \rules) of the retrosynthesis planning space.
\coordinator is also configured to use either LLM-based or value-function based selection,
providing fine-grained control over the retrosynthesis planning process.
\masil uses \verifier whenever a complete route is found, incorporating the natural language feedback from \verifier into \coordinator's context.
While \masil offers fine-grained control over retrosynthesis planning,
it can incur high latency by frequently engaging multi-turn reasoning or defining highly complex value functions.
To partially mitigate this latency issue, \coordinator is configured to follow a special policy for the first $I_{\text{init}}=20$ iterations,
where delegation is skipped and selection follows an efficient single-term value function.
Thus, \masil efficiently explores the retrosynthesis planning space in early iterations,
gathering information to inform fine-grained, agentic decision-making in later iterations.

\subsection{\rfas}
\label{sec:methods:rfas}

\rfas is a \textbf{R}oute-level \textbf{F}eedback \textbf{A}gentic \textbf{S}ystem
that selectively engages \framework components for route-level multi-objective guidance at key decision points during retrosynthesis planning.
\rfas leverages only three agentic components from \framework: \coordinator, \verifier, and \rules.
The \coordinator uses a substantially simplified configuration: only simulation, selection, and expansion are used,
and simulation and selection follow an efficient, static, single-objective value function.
When a route is found, \verifier judges the route.
If the route is rejected, the feedback from the \verifier is provided directly to \rules,
which redefines the boundaries of the search space before planning continues.
Thus, LLM-based agentic reasoning is only performed when a complete route is found, and multi-objective guidance is only provided through \rules and \verifier.
This eliminates many high-latency elements present in \masil (e.g., \steering value function calculations for complex $\valuefn'$, multi-turn reasoning in the delegation step),
providing an efficient, yet still effective, MAS for multi-objective retrosynthesis planning.

\section{Experimental Settings}
\label{sec:experiments}

\subsection{Benchmark tasks}









We carefully curate a benchmark dataset of 218 tasks to evaluate multi-objective retrosynthesis planning methods. 
Each task consists of a target molecule (product) and a set of instructions,
which can express either explicit, hard constraints in \cmo tasks or soft-constraint goals in \gmo tasks. 
Table~\ref{tab:datastats} shows the overall composition of the dataset.
Among the 218 tasks, 111 are \cmo tasks, and 107 are \gmo tasks.
Following \larc, the \cmo tasks consider three types of safety-related constraints: 
\textbf{(1)} avoiding carcinogens, \textbf{(2)} avoiding pyrophoric substances, and \textbf{(3)} avoiding user-specified substances.
Among the 111 \cmo tasks, 48 are single-constraint \cmo tasks, 
sourced from \larc's benchmark set; the other 63 are new, multi-constraint \cmo tasks. 

\begin{table}[]
\centering
\begin{threeparttable}
\caption{Benchmark dataset statistics}
\label{tab:datastats}
\begin{tabular}{p{0.20\linewidth} p{0.25\linewidth} >{\raggedleft\arraybackslash}p{0.25\linewidth}}
\toprule
\makecell[l]{Task Type} & \makecell[l]{Constraint Type} & \makecell[r]{\# Tasks} \\
\midrule
\multirow{7}{*}{\cmo} & C & 28 \\
 & P & 12 \\
 & U & 8 \\
 & C \& P & 3 \\
 & C \& U & 45 \\
 & P \& U & 7 \\
 & C \& P \& U & 8 \\[5pt]
\multirow{1}{*}{\gmo} & $\downarrow$All & 107 \\
\bottomrule
\end{tabular}

\begin{tablenotes}[flushleft]
\footnotesize
\item C, P, and U denote hard constraints of avoiding carcinogenic, pyrophoric, and user-specified substances, respectively.
\& indicates that multiple constraints must be satisfied.
$\downarrow$All indicates that \gmo tasks optimize for all available objectives: \carcinogenicity, \pyrophoricity, \ghs, \totalcost, and \routelength
\end{tablenotes}
\end{threeparttable}
\end{table}

These 63 tasks were identified based on the results of a general-purpose, single-objective retrosynthesis 
planner, \meea, on products from USPTO-190~\citep{meea, retrostar} and BBBP~\citep{meea} test sets, following 
\larc's curation protocol---tasks were selected when \meea could produce a valid route for a product 
that violated \emph{two or more types} of constraints. 
The 107 {\gmo} tasks are each defined by a unique product and a common instruction to plan the best possible route across all safety and cost objectives measured by {\tools}.
Products for the {\gmo} tasks are taken from the {\cmo} tasks, excluding duplicates.
The task details are provided in the SI\dag. 
%

%
%


\subsection{Baselines}
We compare against different baselines on \cmo and \gmo retrosynthesis planning tasks.
For \cmo tasks,
we evaluate two types of baselines:
\textbf{(1)} \larc~\citep{larc}, a state-of-the-art Agent-as-a-Judge for \cmo tasks; and 
\textbf{(2)} general-purpose LLMs, including \claude~\citep{claude4_5}, \gpt~\citep{gpt5_1}, \qwenEightyB~\citep{qwen80b},
\kimi~\citep{kimi}, and \deepseek~\citep{deepseek}.
To ensure a fair comparison, \larc was implemented using the same base models as \masil and \rfas.
All other settings for \larc, including prompts, tools, and hyperparameters, were taken from its original publication~\citep{larc}.
General-purpose LLMs were chosen as representative models of their respective types (e.g., open-weight, closed-source, reasoning-mode, instruct-mode)
with demonstrated strong performance on many tasks~\citep{gpt5_1,deepseek,claude4_5,kimi,qwen80b}.
These LLMs serve as minimal baselines which lack the iterative planning algorithms or tooling present in other baselines methods,
but can still attempt multi-objective retrosynthesis tasks.

For \gmo tasks, we evaluate the same general-purpose LLMs, along with two additional baselines:
\textbf{(1)} \static, a general-purpose retrosynthesis planner subject to a fixed set of multi-objective molecule and reaction restrictions determined at the start of planning based on the target molecule (product) structure;
and
\textbf{(2)} \pareto, a post-processing baseline that measures the average performance of the Pareto-optimal subset of the first 10 routes generated by a general-purpose, single-objective planner.
Please note, \larc cannot be evaluated on \gmo tasks because it requires an explicit constraint as input.

\static, \pareto, \larc, \masil, and \rfas leverage the same underlying \meea planner.
This ensures fairness among non-LLM methods on the benchmark tasks, 
which were designed to challenge \meea-based methods on multi-objective retrosynthesis planning.
Furthermore, \static, \pareto, \masil, and \rfas leverage the exact same underlying tools from \tools in their construction, and \larc shares a very similar set of tools.
Thus, the differences in performance for these methods can primarily be attributed to the methods themselves, not \tools.
Implementation details for the baselines are provided in the SI\dag. 

%




\subsection{Metrics}
%
Following existing work~\citep{larc}, 
%
%
we define three metrics for multi-objective retrosynthesis planning tasks:
\textbf{(1)} presence rate (\presence), which requires routes to be non-empty, 
\textbf{(2)} validity rate (\validity), which requires routes to be meaningful sets of chemical reactions connecting the target molecule to purchasable starting materials,
and \textbf{(3)} success rate (\success), which requires routes to be valid and satisfy any explicitly specified constraints.
We further define additional five metrics related to safety and cost for each route, computed using \tools.
Safety metrics include
\textbf{(1)} \carcinogenicity, the maximum predicted carcinogenicity score for molecules in the route;
\textbf{(2)} \pyrophoricity, a binary value indicating whether the route uses pyrophoric materials;
and
\textbf{(3)} \ghs, the number of globally harmonized system (GHS) hazard statements,
for all molecules in the route.
Cost metrics include \textbf{(4)} \totalcost, the sum of gram-quantity price estimates for all route starting materials;
and \textbf{(5)} \routelength, the number of reactions in the retrosynthesis route.
While \routelength is cost-related, it can also serve as an overall measure of route quality: it is frequently used as such in retrosynthesis planning literature~\citep{retrostar, meea, sirp}; shorter routes are generally preferred.
For fairness, all methods are evaluated under a two-hour time limit and an iteration limit of 500---failure to return a route within these limits is treated as returning an empty route.
Please note, this time limit is inspired by the human retrosynthesis planning times reported in \larc~\citep{larc},
where human retrosynthesis planners took approximately one hour on average to complete single-constraint \cmo tasks.
The two-hour limit provides more time to accommodate more complex tasks with additional constraints, while serving as a practical and meaningful cutoff for multi-objective retrosynthesis planning.

In our experiments, we evaluate \cmo tasks on \presence, \validity, \success, and average \routelength,
using \success to measure the safety and cost profile relative to the explicit constraint.
For \gmo tasks, we report averages across all valid routes for
\carcinogenicity, \pyrophoricity, \ghs, \totalcost, and \routelength,
as well as \presence and \validity.
Pareto dominance~\citep{pareto} is also used to measure route quality in the multi-objective space defined by the safety and cost metrics,
comparing pairs of valid routes from different methods on the same \gmo tasks.
Please note, our evaluation is intentionally focused on objectives that \tools can measure and guide.
Our metrics align with this goal, measuring whether generated routes meet \tools-based definitions of route quality.
We also examine feasibility of reactions in generated routes, with metrics and results discussed in Section~\ref{sec:results:feasibility}
%

\section{Results}
\label{sec:results}

\subsection{\cmo Tasks}
Table~\ref{tab:constrained} shows that \rfas using \qwen can achieve the best \success
of 48.6\% on \cmo tasks,
outperforming \larc and the LLM baselines.
\rfas achieves this by enforcing hard constraints on the retrosynthesis planning space using the regulations generated by \rules and route-level judgments from \verifier.
Meanwhile, \larc, the best baseline, achieves a slightly lower \success (47.7\%) using \mistral than \rfas through soft, value-function-based guidance driven by reaction-level agentic judgments.
While both approaches can produce similar overall \success, 
\rfas achieves higher \success on multi-constraint \cmo tasks (38.1\%), 
noticably higher than \larc's best \success (28.6\%).
This suggests that \rfas's  approach is better-suited to the challenging, multi-constraint \cmo tasks,
while \larc may be sufficient for simpler, single-constraint \cmo settings.
Overall, these results demonstrate that \rfas, an agentic system built with \framework, 
is especially useful on the challenging, multi-constraint \cmo tasks.

\begin{table*}[t]
\setlength{\tabcolsep}{3pt}
\centering
\begin{threeparttable}
\caption{Performance comparison on \cmo tasks
}
\label{tab:constrained}
\begin{tabular}{
   @{\hspace{0pt}}l@{\hspace{2pt}}
   @{\hspace{2pt}}l@{\hspace{5pt}}
    %
    >{\raggedleft\arraybackslash}p{0.04\linewidth}
    >{\raggedleft\arraybackslash}p{0.04\linewidth}
    >{\raggedleft\arraybackslash}p{0.055\linewidth}
    >{\raggedleft\arraybackslash}p{0.04\linewidth}
    >{\raggedleft\arraybackslash}p{0.04\linewidth}
    >{\raggedleft\arraybackslash}p{0.055\linewidth}
    @{\hspace{5pt}}c@{\hspace{5pt}}    
    >{\raggedleft\arraybackslash}p{0.05\linewidth}
    >{\raggedleft\arraybackslash}p{0.05\linewidth}
    >{\raggedleft\arraybackslash}p{0.055\linewidth}
    @{\hspace{2pt}}c@{\hspace{5pt}}    
    >{\raggedleft\arraybackslash}p{0.05\linewidth}
    >{\raggedleft\arraybackslash}p{0.05\linewidth}
    >{\raggedleft\arraybackslash}p{0.055\linewidth}
}
\toprule
 &  & \multicolumn{6}{c}{All Constrained Tasks} && \multicolumn{3}{c}{Multi-constraint} && \multicolumn{3}{c}{Single-constraint} \\ 
 \cmidrule{3-8}
 \cmidrule{10-12}
 \cmidrule{14-16} 
Method & Base Model 
& $\uparrow$\success (\%) & $\uparrow$\validity (\%) & $\uparrow$\presence (\%) & $\downarrow$\ps &
$\uparrow$\SP (\%) & 
$\downarrow$\routelength & 
& $\uparrow$\success (\%) & $\uparrow$\validity (\%) & $\uparrow$\presence (\%) & 
& $\uparrow$\success (\%) & $\uparrow$\validity (\%) & $\uparrow$\presence (\%) \\
\midrule

\multirow{3}{*}{\rfas} & {\textsc{Qwen30B}}
& \textbf{48.6} & 64.0 & 64.0 & 15.3 & 75.9& 6.69 &
& \underline{31.7} & 52.4 & 52.4 & 
& \underline{70.8} & \underline{79.2} & 79.2 \\

 & {\textsc{Mistral}}
& 21.6 & 35.1 & 35.1 & 13.5 & 61.5& 7.05 & 
& 11.1 & 30.2 & 30.2 & 
& 35.4 & 41.7 & 41.7 \\

 & {\textsc{Claude}}
& 45.9 & 66.7 & 66.7 & 20.7 & 68.8& 7.14 & 
& \textbf{38.1} & \textbf{58.7} & 58.7 & 
& 56.3 & 77.1 & 77.1 \\
\cmidrule{2-16}

\multirow{3}{*}{\masil} & {\textsc{Qwen30B}}
& 18.9 & 27.9 & 27.9 & 9.0 & 67.7& \underline{3.48} & 
& 6.3 & 9.5 & 9.5 & 
& 35.4 & 52.1 & 52.1 \\

 & {\textsc{Mistral}}
& 12.6 & 26.1 & 26.1 & 13.5 & 48.3& 3.72 & 
& 0.0 & 12.7 & 12.7 & 
& 29.2 & 43.8 & 43.8 \\

 & {\textsc{Claude}}
& 33.3 & 38.7 & 38.7 & \textbf{5.4} & \textbf{86.0}& 4.12 & 
& 19.0 & 22.2 & 22.2 & 
& 52.1 & 60.4 & 60.4 \\
\cmidrule{2-16}

\multirow{3}{*}{\larc} & {\textsc{Qwen30B}}
& 38.7 & \textbf{72.1} & \textbf{72.1} & 33.3 & 53.7& 5.91 & 
& 23.8 & \textbf{58.7} & 58.7 & 
& 58.3 & \textbf{89.6} & 89.6 \\

 & {\textsc{Mistral}}
& \underline{47.7} & \underline{71.2} & \underline{71.2} & 23.4 & 67.0& 6.67 & 
& 28.6 & \underline{57.1} & 57.1 & 
& \textbf{72.9} & \textbf{89.6} & 89.6 \\

 & {\textsc{Claude}}
& 27.0 & 35.1 & 35.1 & \underline{8.1} & \underline{76.9}& 4.54 & 
& 11.1 & 19.0 & 19.0 & 
& 47.9 & 56.3 & 56.3 \\

\midrule

\multicolumn{2}{l}{{\textsc{Claude}}} 
& 15.3 & 17.1 & \underline{99.1} & 83.8 & 15.4& 3.69 & 
& 17.5 & 17.5 & \underline{98.4} & 
& 12.5 & 16.7 & \textbf{100.0} \\

\multicolumn{2}{l}{\gpt}  
& 9.0 & 15.3 & 95.5 & 86.5 & 9.4& \textbf{2.73} & 
& 9.5 & 17.5 & 92.1 & 
& 8.3 & 12.5 & \textbf{100.0} \\

\multicolumn{2}{l}{{\textsc{Qwen80B}}}
& 7.2 & 9.0 & 98.2 & 91.0 & 7.3& 3.92 & 
& 6.3 & 7.9 & \underline{98.4} & 
& 8.3 & 10.4 & \underline{97.9} \\

\multicolumn{2}{l}{\kimi} 
& 10.8 & 16.2 & 94.6 & 83.8 & 11.4& 4.87 & 
& 9.5 & 15.9 & 90.5 & 
& 12.5 & 16.7 & \textbf{100.0} \\

\multicolumn{2}{l}{\deepseek} 
& 7.2 & 15.3 & \textbf{100.0} & 92.8 & 7.2& 5.05 & 
& 9.5 & 17.5 & \textbf{100.0} & 
& 4.2 & 12.5 & \textbf{100.0} \\

\bottomrule
\end{tabular}
\begin{tablenotes}[flushleft] 
\footnotesize
\item\hspace{-6pt}
\textsc{Qwen30B}: Qwen3 30B Instruct; \textsc{Mistral}: Mistral Nemo 12B; \textsc{Claude}: Claude 4.5 Sonnet; 
\textsc{Qwen80B}: Qwen3 Next 80B Instruct. 
\ps: $\presence - \success$, the difference between \presence and \success; \SP: the ratio of \success to \presence;
Best performance on each metric is in \textbf{bold}, and second-best performance is \underline{underlined}.
\end{tablenotes}
\end{threeparttable}
\end{table*}

\masil achieves a \success of at most 33.3\% (using \claude),
which is lower than the best success rates of \rfas (48.6\%) and \larc (47.7\%).
This reflects \masil's complex design, which results in higher retrosynthesis planning latency and fewer 
tasks completed within the 2-hour time limit---its maximum presence rate is only 38.7\% (\claude).
However, \masil still demonstrates an interesting result: it achieves the highest \success to \presence ratio (86.0\%using \claude) of all methods.
This indicates that, given a sufficiently capable base LLM,
\masil can diligently enforce explicit constraints, prioritizing quality of returned routes over the quantity of tasks completed.
%
When comparing agentic methods within the same base model, \masil has the lowest differences between \presence and \success.
This demonstrates \masil's conservative nature:
it consistently returns the lowest absolute number of routes that violate constraints,
even if it also returns the lowest number of routes.
Furthermore, the routes returned by \masil are shorter on average than those returned by other agentic methods.
This suggests that \masil is better at balancing the objective of explicit constraint satisfaction with the implicit objective of short route length.
Please note, this is not an artifact of selection bias towards less complex synthesis tasks with short routes.
For tasks with non-empty results from
both \masil and the next best agentic method for \routelength, \larc,
\masil produces a route of shorter or equal length in 90.0\% of tasks, with strictly shorter routes in 48.3\%.
These results indicate that \masil's strengths lie in its diligent and thorough attention to route quality,
not in its efficiency.

The LLM baselines perform poorly overall on \cmo retrosynthesis planning, with the best LLM baseline (\claude) achieving only a 15.3\% \success.
Unlike \masil, \rfas, or \larc, these LLM baselines lack specialized training or tooling for retrosynthesis planning,
resulting in low \validity and ultimately low \success.
Interestingly, some LLM baselines achieve relatively low average \routelength (2.73 for \gpt, 3.69 for \claude).
However, the LLM-generated reactions in these routes are not always
realistic chemical transformations, as discussed in Section~\ref{sec:results:feasibility}.
In contrast, \masil, \rfas, and \larc all use the same, specialized method for reaction prediction,
providing meaningful reactions grounded in chemistry principles.
These results highlight the power of \framework, which combines the strengths of LLM-based reasoning, chemistry-specific tools, and retrosynthesis planning algorithms
to produce multi-agent, multi-objective retrosynthesis planning systems with different strengths.
Interestingly, each \framework method peaks in performance under a different base model:
\rfas performs best using \qwen, whereas \masil performs best using \claude.
This indicates that \framework systems may require different capability levels depending on their configuration,
even though they share some of the same components and prompt patterns.
The \success of \rfas increases dramatically from \mistral to \qwen, then decreases only slightly from \qwen to \claude.
This suggests that the simplified design of \rfas may be well-suited to medium-sized models, such as \qwen, and larger models may not provide much additional benefit.
In contrast, \masil's \success consistently increases with model size,
suggesting it requires strong capabilities of the base model.
Intuitively, this follows from the complex design of \masil,
which relies heavily on LLM-based reasoning throughout retrosynthesis planning.
While this allows LLMs substantial freedom to define and refine the guidance and boundaries of the retrosynthesis planning process,
it also requires a base model with very strong chemistry reasoning capabilities to perform well.
This also suggests that \masil may not yet have achieved its best performance under currently available base models,
providing opportunities for future improvement as base models continue to advance.
Overall, these results illustrate how different \framework system designs can benefit from different model capabilities.

%


%
%

\subsection{\gmo Tasks}
Table~\ref{tab:multiobj} shows that {among \masil, \rfas, \static and \pareto, \masil with \qwen excels on 
\gmo tasks, producing routes that have the best average metrics on all safety and cost 
objectives.}
{\rfas and \static are comparably the second-best methods, relying on rules to restrict undesirable reactions and molecules to incorporate safety and cost objectives into retrosynthesis planning.}
While \masil can also impose such rules, it can also directly incentivize desirable reactions and molecules using \steering's value-function-based guidance.
These results highlight the utility of \framework's \steering component,
illustrating how it can guide the retrosynthesis planning towards higher quality routes.

\begin{table*}[t]
\centering
\setlength{\tabcolsep}{3pt}
\begin{threeparttable}
\caption{Performance comparison on \gmo tasks}
\label{tab:multiobj}
\begin{tabular}{
    @{\hspace{5pt}}l@{\hspace{5pt}}
    @{\hspace{5pt}}l@{\hspace{5pt}}
    @{\hspace{5pt}}c@{\hspace{5pt}}
    >{\raggedleft\arraybackslash}p{0.07\linewidth}
    >{\raggedleft\arraybackslash}p{0.07\linewidth}
    >{\raggedleft\arraybackslash}p{0.06\linewidth}
    >{\raggedleft\arraybackslash}p{0.08\linewidth}
    >{\raggedleft\arraybackslash}p{0.05\linewidth}
    @{\hspace{5pt}}c@{\hspace{5pt}}
    >{\raggedleft\arraybackslash}p{0.1\linewidth}
    >{\raggedleft\arraybackslash}p{0.1\linewidth}
}
\toprule
Method & Base Model &
& $\downarrow$\carcinogenicity
& $\downarrow$\pyrophoricity
& $\downarrow$\ghs
& $\downarrow$\totalcost
& $\downarrow$\routelength &
& $\uparrow$\presence (\%)
& $\uparrow$\validity (\%)\\ 
\midrule

\multirow{3}{*}{\masil}
& {\textsc{Qwen30B}}
& 
& 0.670 & \textbf{0.000} & \underline{3.71} & 392.14 & \underline{3.08} 
& 
& 22.4 & 22.4 \\

& {\textsc{Mistral}}
& 
& 0.744 & 0.333 & 4.67 & 419.57 & 3.67 
& 
& 5.6 & 5.6 \\

& {\textsc{Claude}}
& 
& 0.709 & 0.088 & 4.12 & 427.27 & 3.29 
& 
& 31.8 & 31.8 \\
\cmidrule{2-11}

\multirow{3}{*}{\rfas}
& {\textsc{Qwen30B}}
& 
& 0.756 & 0.054 & 5.91 & 705.28 & 5.75 
& 
& \underline{52.3} & \underline{52.3} \\

& {\textsc{Mistral}}
& 
& 0.732 & 0.102 & 5.53 & 643.22 & 6.10 
& 
& 45.8 & 45.8 \\

& {\textsc{Claude}}
& 
& 0.750 & 0.086 & 5.83 & 687.99 & 6.06 
& 
& \textbf{65.4} & \textbf{65.4} \\
\midrule

\static &
& 
& 0.748 & 0.070 & 6.26 & 627.44 & 7.40 
& 
& 53.3 & 53.3 \\

\pareto &
& 
& 0.788 & 0.110 & 6.64 & 701.18 & 8.90 
& 
& \textbf{96.3} & \textbf{96.3} \\

\midrule

\multicolumn{2}{l}{{\textsc{Claude}}}
& 
& 0.670 & \underline{0.033} & 4.67 & 232.85 & 3.87 
& 
& \textbf{100.0} & 28.0 \\

\multicolumn{2}{l}{\gpt}
& 
& \textbf{0.621} & 0.200 & 4.60 & \textbf{73.18} & 3.40 
& 
& \textbf{100.0} & 4.7 \\

\multicolumn{2}{l}{{\textsc{Qwen80B}}}
& 
& \underline{0.626} & 0.105 & \textbf{3.32} & \underline{143.44} & \textbf{3.00} 
& 
& 98.1 & 17.8 \\

\multicolumn{2}{l}{\kimi}
& 
& 0.766 & 0.500 & 7.90 & 658.81 & 6.40 
& 
& \underline{99.1} & 18.7 \\

\multicolumn{2}{l}{\deepseek}
& 
& 0.700 & 0.222 & 4.33 & 172.61 & 3.78 
& 
& 98.1 & 8.4 \\
\bottomrule
\end{tabular}
\begin{tablenotes}[flushleft]
\footnotesize 
\item 
Abbreviations and formatting follow the conventions used in Table~\ref{tab:constrained}.
\end{tablenotes}
\end{threeparttable}
\end{table*}

We also use Pareto-dominance to evaluate \masil routes, showing that they are consistently better than or comparable to \static routes from a holistic, multi-objective perspective.
We use Pareto-dominance to compare \masil (using \claude as the base model) and 
\static on the 22 tasks completed by both methods.
\masil's route Pareto-dominates the \static route in 8 (36.4\%) of tasks; in the remaining tasks, neither method dominates.
This indicates that \masil routes are consistently better than or comparable to \static routes from a holistic, multi-objective perspective.
A similar pattern emerges when comparing \masil and \rfas on the 29 tasks they both complete using \claude as the base model.
\masil routes dominate the \rfas routes in 10 tasks (34.4\%),
neither method dominates in 15 tasks (51.7\%),
and \rfas dominates \masil in only 4 tasks (13.8\%).
These results highlight \masil's ability to plan higher-quality routes than other methods from a multi-objective perspective,
demonstrating that \framework can be used to construct quality-focused MAS for multi-objective retrosynthesis planning.

\rfas also shows substantial promise on \gmo tasks,
outperforming both \static and \pareto in terms of average \ghs and \routelength while maintaining similar quality for other metrics.
This trend is consistent across base models, suggesting that this benefit is driven by \framework components in the \rfas system,
not simply the capabilities of the base models.
While \rfas and \static use similar, rule-based logic to incorporate safety and cost objectives,
\rfas applies these rules adaptively,
using agentic reasoning to iteratively refine its rules throughout retrosynthesis planning.
In addition,
\rfas also achieves a higher \presence (65.4\%) when using \claude as a base model than \static (53.3\%).
This indicates that the adaptive, agentic capabilities offered by \framework components not only lead to better solutions,
but, in some cases, allow more solutions to be found.
%
%

The LLM baselines appear to perform well on \gmo retrosynthesis planning tasks,
with \claude and \qwenEightyB achieving comparable performance to \masil in terms of average safety and cost metrics as well as \validity.
However, closer inspection of the LLM-generated routes reveals unrealistic reaction 
patterns that may skew these metrics.
For example, some of \claude's valid routes involve reactions in which the products contain atom types not present in the reactants.
Such reactions are chemically impossible in this state,
and any alterations would inherently change the safety and cost profile of the route.
This highlights the importance of \framework's principled approach to reaction prediction,
relying on specialized chemistry models to
produce more chemically reasonable reactions.
Reaction feasibility is examined in greater detail in Section~\ref{sec:results:feasibility}.
\gpt achieves the best \carcinogenicity and \totalcost, but produces valid routes on only 5 tasks (4.7\%), substantially limiting its practical utility.
Please note, this low \validity is paired with 100.0\% \presence,
indicating that \gpt generates routes for most tasks, but nearly all fail to meet basic validity criteria.
This pattern of high \presence and low \validity can be seen across all LLM baselines.
In contrast, \framework systems (e.g., \masil, \rfas) can only generate valid or empty routes.
These results underscore the importance of \framework's specialized design,
allowing \masil and \rfas to provide meaningful retrosynthesis routes with strong safety and cost profiles.
Compared with \masil, \rfas, and \static, \pareto has the worst average metrics on three of five safety and cost objectives.
\pareto is limited to post-hoc filtering of the top-10 routes generated by a planner with no multi-objective guidance,
preventing it from identifying higher-quality routes.
This highlights an important feature of \framework:
the ability to incorporate multi-objective guidance into the process of retrosynthesis planning, not as an afterthought.
Taken together, the results of \pareto, \static, \rfas, and \masil illustrate that as multi-objective guidance becomes more tightly woven into each retrosynthesis planning method,
the generated routes improve in overall quality.

%


\subsection{Reaction Feasibility}
\label{sec:results:feasibility}

\begin{table}[tb]
\centering
\caption{Estimated feasibility of reactions in generated routes}
\label{tab:feasibility}
\begin{threeparttable}
\begin{tabular}{
    p{0.37\linewidth}
    >{\raggedleft\arraybackslash}p{0.13\linewidth}
    >{\raggedleft\arraybackslash}p{0.13\linewidth}
}
\toprule
Method & $\feasibility_{MT}$ (\%) & $\feasibility_{C}$ (\%) \\
\midrule
\masil (\qwen) & \textbf{45.6} & \textbf{69.8} \\
\rfas (\qwen)   & \ul{42.0} & \ul{62.7} \\
\midrule
\claude         & 19.6 & 40.8 \\
\gpt            & 10.6 & 30.6 \\
\qwenEightyB    & 13.0 & 30.4 \\
\kimi           & 31.9 & 56.3 \\
\deepseek       & 26.6 & 48.4 \\
\bottomrule
\end{tabular}
\begin{tablenotes}[flushleft]
\footnotesize
\item \hspace{-3pt}\textbf{Bold} indicates the best performance for each metric; underline indicates the second-best performance.
\end{tablenotes}
\end{threeparttable}
\\
\end{table}

Across all tasks, we analyze 
the feasibility of reactions
across generated routes for all tasks
from \rfas, \masil, and LLM baselines.
Accurately predicting reaction feasibility is highly non-trivial~\citep{retrogfn, rlsync}---to increase confidence in our analysis, we provide two predicted metrics for reaction feasibility:
$\feasibility_{MT}$ and $\feasibility_{C}$.
$\feasibility_{MT}$ uses an independently-trained forward synthesis prediction model to determine reaction feasibility, deciding a reaction is feasible if the product appears in the model's top 5 predicted products for the reactants.
$\feasibility_{C}$ uses \claude as an LLM-as-a-Judge to determine reaction feasibility.

Table~\ref{tab:feasibility} shows that \masil has the highest $\feasibility_{MT}$ (45.6\%) and $\feasibility_{C}$ (69.8\%)
across all reactions in valid routes for all benchmark tasks.
\rfas has the second highest scores at 42.0\% and 62.7\%, respectively---significantly higher than the best LLM baseline, \kimi, which achieved 31.9\% (p=0.006, two-sided, two-sample t-test) and 56.3\% (p=0.005), respectively.
This suggests that the reaction prediction method
used by \masil and \rfas during retrosynthesis planning provides a significant benefit,
ensuring every reaction is generated using patterns learned from real, patented chemical reactions~\citep{retrostar, meea, lowe_chemical_2017}.
In contrast,
LLM baselines directly generate reactions as sequences of tokens,
relying on their internal knowledge and reasoning abilities to ensure chemical transformations are valid.
Among the LLM baselines, \kimi and \deepseek, which were run in reasoning mode, have the highest $\feasibility_{MT}$ and $\feasibility_{C}$,
suggesting that reasoning behaviors provide a substantial benefit to LLM-based reaction generation.
Notably, $\feasibility_{MT}$ and $\feasibility_{C}$ show the same relative ordering of methods, and agree on 68.0\% reactions.
This suggests the comparisons between models reflect real underlying trends,
not simply artifacts of a single, imperfect evaluation metric.
Overall, these results highlight the importance of grounded, principled reaction prediction in retrosynthesis,
improving the probability of generating chemically meaningful retrosynthesis routes.

\subsection{Case Study}

\begin{figure}[htb]
    \centering
    \includegraphics[width=\columnwidth]{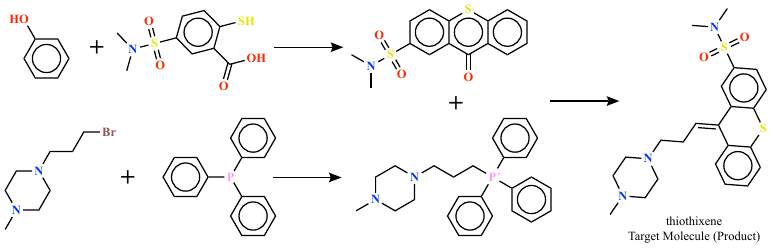}
    \caption{Synthesis of thiothixene generated by \masil for a \gmo task.}
    \label{fig:case}
\end{figure}

Figure~\ref{fig:case} shows a 3-step retrosynthesis route generated by \masil (using \claude as the base model) for a \gmo task with the target molecule thiothixene, an antipsychotic pharmaceutical for the treatment of schizophrenia.
This convergent synthesis route unites two key intermediates,
each prepared from commercially available starting materials, via a final Wittig olefination.
This convergent strategy offers a substantially safer and more efficient alternative to traditional linear routes~\cite{muren_thioxanthene_2002, hobbs_metabolism_1968} by minimizing multi-step handling and hazardous intermediate accumulation.

MASIL's generated route has a strong safety and cost profile, with a maximum \carcinogenicity score of 0.512, no pyrophoric materials, 5 \ghs hazard statements, and a \totalcost of \$269.62.
\masil used all four \framework components to plan this route, providing a narrative of how each component contributes to \masil's success.
During its first 20 iterations without delegation, \masil uncovered a route with 8 steps, \totalcost over \$1,000, and high \carcinogenicity, which was summarily rejected by the \verifier.
Subsequently, \coordinator delegated to \rules to restrict specific molecules with high predicted carcinogenicity,
followed by delegation to \steering to penalize expensive starting materials, deep routes, and carcinogenic substances.
\coordinator continued to leverage \steering and \rules to iteratively adjust its retrosynthesis planning strategy over many iterations.
Occasionally, \coordinator directly used its own reasoning to select specific synthetic directions for expansion, rather than relying on the value function.
Furthermore, \coordinator found an interesting use for \rules beyond restricting safety or cost concerns:
\rules could enhance exploration by restricting common intermediates or reactions when simulation candidates became too similar.
After 49 total iterations, the route shown above was identified.
\verifier accepted the route as a substantial improvement over the first route, and planning terminated.
This illustrates how \framework components work together in multi-objective retrosynthesis planning,
planning routes with substantially better safety and cost profiles.
We further illustrate the impact of \framework components within \masil using ablation studies in Section~\ref{sec:results:ablation}.

\subsection{Ablation Study}
\label{sec:results:ablation}

\begin{table*}[t]
\centering
\setlength{\tabcolsep}{3pt}
\begin{threeparttable}
\caption{Ablation study on \cmo and \gmo tasks using \qwen}
\label{tab:ablation_qwen}
\begin{tabular}{
    @{\hspace{5pt}}l@{\hspace{5pt}}
    @{\hspace{5pt}}c@{\hspace{5pt}}
    >{\raggedleft\arraybackslash}p{0.05\linewidth}
    >{\raggedleft\arraybackslash}p{0.05\linewidth}
    >{\raggedleft\arraybackslash}p{0.05\linewidth}
    @{\hspace{5pt}}c@{\hspace{5pt}}
    >{\raggedleft\arraybackslash}p{0.07\linewidth}
    >{\raggedleft\arraybackslash}p{0.07\linewidth}
    >{\raggedleft\arraybackslash}p{0.06\linewidth}
    >{\raggedleft\arraybackslash}p{0.08\linewidth}
    >{\raggedleft\arraybackslash}p{0.05\linewidth}
    >{\raggedleft\arraybackslash}p{0.05\linewidth}
}
\toprule
Method
&
& \multicolumn{3}{c}{\cmo Tasks}
&
& \multicolumn{6}{c}{\gmo Tasks} \\
\cmidrule{3-5}
\cmidrule{7-12}
&
& $\uparrow$\presence (\%)
& $\uparrow$\success (\%)
& $\downarrow$\routelength
&
& $\downarrow$\carcinogenicity
& $\downarrow$\pyrophoricity
& $\downarrow$\ghs
& $\downarrow$\totalcost
& $\downarrow$\routelength
& $\uparrow$\presence (\%)\\ 
\midrule

\masil
&
& 27.9 & \textbf{18.9} & \textbf{3.48}
&
& \textbf{0.670} & \textbf{0.000} & \textbf{3.71} & \textbf{392.14} & \textbf{3.08} & 22.4 \\

\masil w/o \steering
&
& 27.0 & 18.0 & 4.10
&
& 0.718 & 0.033 & 4.97 & 444.20 & 3.97 & 28.0 \\

\masil w/o \rules
&
& 22.5 & 13.5 & 3.86
&
& 0.748 & 0.045 & 4.82 & 455.68 & 3.86 & 20.6 \\

\masil w/o \verifier
&
& \textbf{43.2} & \textbf{18.9} & 3.81
&
& 0.757 & 0.174 & 5.02 & 646.87 & 3.89 & \textbf{43.0} \\

\bottomrule
\end{tabular}
\begin{tablenotes}[flushleft]
\footnotesize 
\item 
All rows use \qwen as the base model.  The best number in each column is in \textbf{bold}.  \validity has been omitted for brevity; it is equal to \presence for \masil and all ablations.  \ps and \SP are also omitted, but can be inferred from the \presence and \success above.
\end{tablenotes}
\end{threeparttable}
\end{table*}

Table~\ref{tab:ablation_qwen} shows the \cmo and \gmo performance of ablations of \masil, where each of the optional agentic components of \framework (\steering, \rules, and \verifier) are removed individually to characterize their isolated impact on the system.
\masil using \qwen was selected for this  ablation study due to its use of all \framework components and its strong performance on \gmo tasks.
Overall, \masil performs better than its ablations on most metrics,
indicating that \steering, \rules, and \verifier each contribute to the success of \masil.
Please note, these ablation results are specific to \masil;
other configurations and compositions of \framework components may elicit different behaviors and effects.

\subsubsection{\masil without \steering.}
Removing \steering results in only slightly lower \success on \cmo tasks,
indicating that \steering may not be strictly necessary to ensure constraint satisfaction on \cmo tasks.
However, removing \steering also results in the longest average \routelength on both \cmo and \gmo tasks across ablations.
This suggests that \steering may be used by \masil on \cmo tasks to guide planning towards shorter routes.
For \gmo tasks, removing \steering results in decreased performance on every cost and safety metric compared to \masil.
This suggests that \steering is helpful in guiding retrosynthesis planning towards shorter and more optimal routes overall.
Among ablations, removing \steering produces the smallest impact on \carcinogenicity, \pyrophoricity, and \totalcost for \gmo tasks.
This suggests that \rules and \verifier may already help \masil avoid the worst scoring routes on these metrics,
but adding \steering can help guide \masil towards even better routes.

\subsubsection{\masil without \rules.}
Removing \rules results in the lowest success rate among the ablations on \cmo tasks,
indicating that \rules's regulations are an effective way to enforce hard constraints.
Furthermore, removing \rules from \masil leads to substantially decreased performance on every cost and safety metric in \gmo tasks.
This suggests that \rules also plays a role in \gmo tasks,
removing problematic reactions from the retrosynthesis planning space.
Interestingly, \masil without \rules achieves the lowest \presence in both \gmo and \cmo tasks.
This suggests that \rules plays a key role in \masil's success:
\rules strictly prevents \masil from spending time and iterations on suboptimal directions
by regulating the bounds of the retrosynthesis planning process.

\subsubsection{\masil without \verifier.}
Across both \cmo and \gmo task types, \verifier has a substantial impact on \presence.
This highlights a side-effect of \verifier's role within \masil:
by rejecting suboptimal routes,
\verifier reduces the number of tasks for which \masil can generate a route.
However, removing \verifier does not increase \success on \cmo tasks and worsens performance on \gmo cost and safety metrics.
Moreover, \masil without \verifier achieves the worst average \carcinogenicity, \pyrophoricity, \ghs, and \totalcost on \gmo tasks among all ablations.
This suggests that the verification of generated routes is essential to \masil's performance on both \cmo and \gmo tasks.
Without \verifier checking the quality of routes, \masil frequently returns suboptimal results.

Further ablations removing multiple components are possible in principle,
but difficult to justify while retaining the overall design principles and configuration of \masil.
For example, removing both the \steering and the \rules effectively disables delegation, leaving little to no opportunity for MAS multi-objective guidance in the retrosynthesis planning loop.
A full combinatorial study of all components and configurations possible under \framework is beyond the scope of this work, as discussed in Section~\ref{sec:discussion}.

\subsection{Operational Costs of \framework systems}

\begin{table}[]
\centering
\setlength{\tabcolsep}{3pt}
\begin{threeparttable}
\caption{Operational costs for \framework systems}
\label{tab:costs}
\begin{tabular}{
    @{\hspace{5pt}}l@{\hspace{8pt}}
    >{\raggedleft\arraybackslash}p{0.16\linewidth}
    >{\raggedleft\arraybackslash}p{0.16\linewidth}
}
\toprule
Metric & MASIL & RFAS \\
\midrule
Runtime (min) & 71.153 & 4.343 \\
Input Tokens (MTok) & 1.170 & 0.065 \\
Output Tokens (MTok) & 0.038 & 0.003 \\
Total Tokens (MTok) & 1.208 & 0.068 \\
API Calls & 68.289 & 5.509 \\
\bottomrule
\end{tabular}
\begin{tablenotes}[flushleft]
\footnotesize
    \item\hspace{-5pt} Each row indicates the average cost per benchmark task. Runtime is reported in minutes and token usage is reported in millions of tokens. Here, \masil and \rfas use \claude as a base model.
\end{tablenotes}
\end{threeparttable}
\end{table}

Table~\ref{tab:costs} shows the operational costs of running \masil and \rfas on the most expensive base model, \claude, averaged across all 218 tasks.
Overall, we observe that MASIL is substantially more expensive than RFAS on average across all metrics: runtime, token usage, and
number of LLM API calls.
This reflects the design of \masil, which tightly integrates multi-agent reasoning into the retrosynthesis planning.
Please note, even \masil's higher costs are low when considered within practical contexts for retrosynthesis planning, such as drug development,
in which years of time and billions of dollars are spent to develop a single drug~\citep{blass_basic_2021}.
In our experimental settings,  runtime cannot exceed 120 minutes (2 hours),
bounding computational cost for fairness of comparison and practicality of the benchmarking experiments.
In future work, \masil may benefit from longer time limits; however,
care should be taken to ensure fairness and justify the practicality of such long-running experiments.


\section{Discussion and Conclusions}
\label{sec:discussion}

More MAS can be built using \framework, 
with \masil and \rfas being just two representative systems.
Our results demonstrate that \framework can be used to construct different MAS with distinct benefits and tradeoffs for multi-objective retrosynthesis planning,
suggesting that there may be no universally optimal MAS design across all practical settings.
\masil and \rfas exhibit complementary strengths,
which can be traced to distinct design choices within \framework.
\masil generates routes with the strongest safety and cost profiles on \gmo tasks,
suggesting that tightly integrating \framework components into the retrosynthesis planning process is beneficial for open-ended settings where multi-objective route quality is key.
Meanwhile, \rfas excels at generating successful routes in \cmo tasks,
suggesting that selective engagement of \framework components may be preferred when objectives are well-defined and efficient planning is desired.
Together, these insights motivate future work to
investigate strategies to dynamically design \framework systems that align with the requirements of each multi-objective retrosynthesis planning task.

Evaluating quality, safety, and cost objectives in retrosynthesis planning remains a fundamental challenge,
as current computational \tools still falls behind in accurately representing the complexities of real-world synthetic chemistry.
\masil and \rfas rely on \tools to ground the decision making of their \framework components;
inaccuracies could lead to suboptimal multi-objective guidance.
Additionally, scalable benchmark evaluation relies on accurate \tools, which serve as computable proxies for their respective real-world objectives.
This motivates the development of more robust and task-specific computational tools for synthesis evaluation.
Alternatively, future efforts could explore how \framework components can learn to compensate for \tools inaccuracies through enhanced reasoning or internalized chemistry knowledge.

The cost and safety objectives explored here are only the beginning of practical multi-objective retrosynthesis planning.
There are many additional practical considerations for retrosynthesis planning, such as yield, reaction conditions, racemization risk, mechanistic plausibility, and stereoselectivity.
However, scalable and accurate \emph{in silico} tools for these objectives remain an area of open research~\citep{voinarovska_when_2024, ballard_problem_2019, hansen_prediction_2016},
Furthermore, accurate prediction of such objectives would inherently require more information than is currently available in retrosynthesis planners' generated reactions,
which typically include only the reactants and products without specifically naming catalysts or solvents.
This motivates the development of reaction prediction methods that return this additional information as well as new \tools to deepen the multi-objective retrosynthesis planning space that \framework systems can explore.
Ultimately, expert verifications through literature search and laboratory experiments are the gold standard for evaluating retrosynthesis routes.
Future work should consider how to incorporate such information into \framework for both large-scale evaluation and guidance without entirely sacrificing scalability.

Current benchmarks in multi-objective retrosynthesis planning do not yet capture the idiosyncratic needs of individual chemists and retrosynthesis projects.
The dataset in Section~\ref{sec:experiments} is designed to challenge \framework systems and other \emph{in silico} retrosynthesis planners on meaningful hard and soft constraints,
but its construction may not accurately reflect how chemists will eventually use such systems.
Collecting a set of multi-objective tasks submitted by real-world chemists from various academic and industry laboratories
would be an interesting and non-trivial next step in benchmarking multi-objective systems.
Additionally, future work may consider testing \framework MAS in the wild in the style of popular benchmarks from other domains~\citep{lin_wildbench_2025}.

%


%
%


%

\section*{Conflicts of Interest}

There are no conflicts of interest to declare.


\section*{Data Availability Statement}

Code and datasets for \framework are available at \url{https://github.com/ninglab/MMORF}.
The benchmark dataset can also be found in the SI~\dag. 
The repository also includes synthesis routes generated from \masil and \rfas for products and constraints described in this work. Prompts used in \framework implementations are available in the SI~\dag. 
For reproducibility, the repository includes log files containing the inputs and outputs  to proprietary LLM APIs used in the study,
as well as generation dates.
%
References 3, 6--8, 14--16, and 31--36 are cited in the SI~\dag.


\section*{Acknowledgments}

This project was made possible, in part, by support from the National Science Foundation
grant nos. IIS-2133650 and IIS-2435819, and the National Library of Medicine grant no. 1R01LM014385.
Any opinions, findings, conclusions or recommendations expressed in this
paper are those of the authors and do not necessarily reflect the views of the funding agency.
The authors would like to thank Benjamin Burns for his proofreading of the manuscript.

\bibliography{colm2026_conference}
\bibliographystyle{rsc} 

\onecolumn
\appendix
{\noindent\Huge MMORF: A Multi-agent Framework for Designing Multi-objective Retrosynthesis Planning Systems \\}
{\\\huge Supplementary Information}
{\\[2pt]Frazier N. Baker$^{a}$, Trieu Nguyen$^{a}$, Reza Averly$^{a}$, Botao Yu$^{a}$, Daniel Adu-ampratwum$^{b}$, Huan
Sun$^{a}$, and Xia Ning$^{a,b,c,d}$}

\footnotetext{\textit{$^{a}$~Department of Computer Science and Engineering, The Ohio State University, 395 Dreese Laboratories, 2015 Neil Avenue, Columbus, Ohio, United States of America.}}
\footnotetext{\textit{$^{b}$~Division of Medicinal Chemistry and Pharmacognosy, The Ohio State University, 217 Lloyd M. Parks Hall, 500 West 12th Avenue, Columbus, Ohio, United States of America.}}
\footnotetext{\textit{$^{c}$~Department of Biomedical Informatics, The Ohio State University, 250 Lincoln Tower, 1800 Cannon Drive, Columbus, Ohio, United States of America.}}
\footnotetext{\textit{$^{d}$~Translational Data Analytics Institute, The Ohio State University, 175 Pomerene Hall, 1760 Neil Avenue, Columbus, Ohio, United States of America.}}
\footnotetext{References for the supplementary information are included in the main article.}

\setcounter{page}{1}

\renewcommand{\thetable}{A\arabic{table}}
\renewcommand{\thefigure}{A\arabic{figure}}

\setcounter{table}{0}
\setcounter{figure}{0}
\section{Dataset}
\label{app:dataset}

\subsection{\cmo Tasks}

Table~\ref{tab:dataset} shows the \cmo tasks in the benchmark dataset.
For \cmo tasks, the instruction includes the target molecule (product) SMILES.
In Table~\ref{tab:dataset}, we assign an ID to each task, with the prefix S  and M indicating \cmo tasks with a single constraint and multiple constraints, respectively.
The constraint type(s) are also listed,
with `C', `P', and `U' indicating the task requires avoiding carcinogenic, pyrophoric, and user-specified substances, respectively.

{
\renewcommand{\arraystretch}{1.2}
\begin{longtable}{>{\raggedright\centering}p{0.035\linewidth}
>{\centering\arraybackslash}p{0.10\linewidth}
>{\raggedright\arraybackslash}p{0.76\linewidth}}
\caption{\cmo Tasks}
\label{tab:dataset}\\
\toprule
ID & Constraint Type(s) & Instruction \\
\midrule
\endfirsthead

\toprule
ID & Constraint Type(s) & Instruction \\
\midrule
\endhead

\midrule
\multicolumn{3}{r}{\emph{Continued on next page}}\\
\midrule
\endfoot

\hline
\endlastfoot

S1  &  \mbox{C}     & Find the shortest synthesis route for \texttt{O=S(=O)(C\#Cc1ccc(Cl)cc1)N1CCNCC1} that   avoids all carcinogenic substances.                                                                            \\
S2  &  \mbox{C}     & Identify a synthesis route for   \texttt{CC(C)(C)OC(=O)NC1\allowbreak{}(c2nc(O)c3c(Cl)ccn3n2)\allowbreak{}CC1} that avoids known or suspected   carcinogens.                                                                     \\
S3  &  \mbox{C}     & Plan the synthesis of \texttt{CC(=O)OCc1nc2cnc(Br)cc2n1C(C)(C)COC(C)=O},   avoiding any carcinogenic substances.                                                                                       \\
S4  &  \mbox{C}     & Plan the synthesis of   \texttt{FC(F)(F)Cn1ncnc1-c1cc2n(n1)-c1cc(C3CCNCC3)\allowbreak{}ccc1OCC2}, avoiding any known or suspected carcinogens.                                                                    \\
S5  &  \mbox{C}     & Perform synthesis planning for   \texttt{C=C(C{[}C@@H{]}(Cc1ccc(-c2ccccc2)cc1)\allowbreak{}NC(=O)OC(C)(C)C)C(=O)O} without using any   carcinogenic substances.                                                     \\
S6  &  \mbox{C}     & Provide a synthesis route for   \texttt{COc1cc2c(Oc3cc(C)c(C)nc3-\allowbreak{}c3cccc(C)n3)ccnc2cc1OCCNCCO} that avoids   carcinogens.                                                                               \\
S7  &  \mbox{C}     & Find the shortest synthesis route possible for   \texttt{C\#CC1(O)C(C)\allowbreak{}=CC2(CC1(C)C(F)(F)F)OC(C)C(C)O2} without using carcinogens.                                                                      \\
S8  &  \mbox{C}     & Identify the shortest possible synthetic route for   \texttt{ClCc1ccc2c(c1)Nc1nccnc1S2} that avoids carcinogens.                                                                                       \\
S9  &  \mbox{C}     & Identify the best synthesis route for   \texttt{COC(=O)\allowbreak{}c1ccc2c(c1)C=CC(=C(Cl)Cl)CO2} that avoids carcinogenic materials.                                                                               \\
S10 &  \mbox{C}     & Design a synthesis plan for   \texttt{C{[}C@@H{]}(O)c1nc2cnc3ccsc3c2n1\allowbreak{}{[}C@H{]}1CC{[}C@H{]}\allowbreak{}(CO)CC1} without using any   carcinogens.                                                                   \\
S11 &  \mbox{C}     & Provide the shortest synthesis route for   \texttt{Cn1oc(=O)\allowbreak{}nc1/C(=N\textbackslash{}\textbackslash{}OCc1cccc(N)n1)c1ccccc1} that does not use any   carcinogens.                                       \\
S12 &  \mbox{C}     & Plan the synthesis of \texttt{CC(C)(CO)n1c(CO)nc2cnc(Br)cc21} without using   carcinogens.                                                                                                             \\
S13 &  \mbox{C}     & Plan the synthesis of   \texttt{CCCC{[}Sn{]}(/C=C/C1(O)C(C)=CC2\allowbreak{}(CC1\allowbreak{}(C)C(F)\allowbreak{}(F)F)\allowbreak{}OC(C)C(C)O2)\allowbreak{}(CCCC)CCCC}, but do   not use any carcinogens in your synthesis route.                                      \\
S14 &  \mbox{C}     & Find the shortest synthesis path for   \texttt{COC(=O)CCc1cc2cc\allowbreak{}(-c3noc\allowbreak{}(-c4ccc\allowbreak{}(OC(C)C)c(Cl)c4)n3)ccc2n1C} that does not use   any carcinogenic substances.                                              \\
S15 &  \mbox{C}     & Plan the synthesis of   \texttt{CC(C)(C)OC(=O)N1CC=C(c2ccc3c(c2)-n2nc\allowbreak{}(-c4ncnn4CC(F)(F)F)cc2CCO3)CC1} without   using any carcinogens.                                                                  \\
S16 &  \mbox{C}     & Find the shortest synthesis route for   \texttt{C{[}Si{]}(C)(C)CCOCn1cc\allowbreak{}(C2CCc3c\allowbreak{}(C(=O)O)nn(COCC{[}Si{]}(C)(C)C)\allowbreak{}c3C2)cn1} that doesn't   use any carcinogenic substances.                                \\
S17 &  \mbox{C}     & Plan the synthesis of \texttt{O=C(OCc1ccccc1)N1CC{[}C@H{]}2CCCN\allowbreak{}(CCc3ccccc3)C{[}C@H{]}21}   without using carcinogens.                                                                                  \\
S18 &  \mbox{C}     & Find a synthesis route for   \texttt{Cn1oc(=O)nc1/C(=N\textbackslash{}OCc1cccc\allowbreak{}(NC(=O)\allowbreak{}OCCc2ccccc2)n1)c1ccccc1} that avoids all   known or suspected carcinogens.                        \\
S19 &  \mbox{C}     & Plan the shortest synthesis route for   \texttt{CC(C)(C)OC(=O)NC1\allowbreak{}(c2nc\allowbreak{}(NCc3ccccn3)\allowbreak{}c3c(Cl)ccn3n2)CC1}. Do not use any   carcinogens in the route.                                                       \\
S20 &  \mbox{C}     & Generate a synthesis plan for the following compound without using any   carcinogens: \texttt{CC{[}C@@H{]}(OC(=O)c1ccccc1){[}C@H{]}1CCCN(C(=O)OC(C)(C)C)C1}.                                            \\
S21 &  \mbox{C}     & Plan the synthesis of   \texttt{CC(C)(C)OC(=O)N{[}C@@H{]}1c2cccnc2{[}C@H{]}\allowbreak{}(O)CC{[}C@H{]}1c1cccc(F)c1F}. Avoid   carcinogenic materials.                                                               \\
S22 &  \mbox{C}     & Find the shortest synthesis route that doesn't use carcinogens for the   following product:   \texttt{CC(C)(C)OC(=O)N{[}C@@H{]}1\allowbreak{}c2cccnc2{[}C@H{]}(N)CC{[}C@H{]}1c1cccc(F)c1F}.                         \\
S23 &  \mbox{C}     & Design a synthesis path for   \texttt{CC(C)(C)OC(=O)N{[}C@@H{]}1\allowbreak{}c2cccnc2C(=O)\allowbreak{}CC{[}C@H{]}1c1cccc(F)c1F}. Do not use any   carcinogens.                                                                  \\
S24 &  \mbox{C}     & Plan the shortest possible synthesis route for \texttt{COCCCc1cc\allowbreak{}(CN(C(=O)\allowbreak{}{[}C@H{]}2CNCC{[}C@@H{]}2c2ccc\allowbreak{}(OCCOc3c(Cl)cc\allowbreak{}(C)cc3Cl)\allowbreak{}cc2)\allowbreak{}C2CC2)\allowbreak{}cc\allowbreak{}(OCCOC)c1},   but do not use any carcinogens in the route. \\
S25 &  \mbox{C}     & Find the shortest synthesis path for   \texttt{COCCCc1cc\allowbreak{}(CN(C(=O){[}C@H{]}2CN\allowbreak{}(C(=O)OC(C)(C)C)CC{[}C@@H{]}2\allowbreak{}c2ccc(OCCOc3c(Cl)\allowbreak{}cc(C)\allowbreak{}cc3Cl)\allowbreak{}cc2)\allowbreak{}C2CC2)\allowbreak{}cc(OCCOC)c1}.   Do not use any carcinogens.            \\
S26 &  \mbox{C}     & Generate a synthesis plan without carcinogens for   \texttt{CC\#CCn1c(Br)nc(C=O)c1C(=O)OC}.                                                                                                            \\
S27 &  \mbox{C}     & Find the shortest synthesis path for \texttt{COC(=O)c1ccc2c(c1)C=CC(=CCl)CO2}   that does not use carcinogenic substances.                                                                             \\
S28 &  \mbox{C}     & Find a synthesis path for   \texttt{O=C(Nc1cccc(Cl)c1)N1CCc2{[}nH{]}nc\allowbreak{}(C(=O)N3CC(F)CO3)c2C1} that doesn't use   carcinogens.                            \\
\midrule
S29  &  \mbox{P}     & Identify a synthesis route for   \texttt{C{[}C@H{]}(O{[}Si{]}(C)(C)C(C)(C)C){[}C@@H{]}1CC(=O)CC(C)(C)N1} that does not use   pyrophoric substances.                                                    \\
S30  &  \mbox{P}     & Find the shortest synthesis plan for   \texttt{C{[}C@H{]}(c1ccccc1)\allowbreak{}N1C{[}C@{]}2\allowbreak{}(C(=O)\allowbreak{}OC(C)(C)\allowbreak{}C)C=CC{[}C@@H{]}2C1\allowbreak{}=S} that avoids all   pyrophoric and water-reactive substances.                        \\
S31  &  \mbox{P}     & Plan the shortest synthesis route for \texttt{CC(=O)c1ccc2c(c1)C=CC(O)(CO)CO2}   without using any pyrophoric or water reactive substances.                                                            \\
S32  &  \mbox{P}     & Synthesize \texttt{CC{[}C@@H{]}(OC(=O)c1ccccc1){[}C@H{]}1CCCN(C(=O)OC(C)(C)C)C1} without   using any pyrophoric or water-reactive substances.                                                          \\
S33  &  \mbox{P}     & Perform synthesis planning for \texttt{CC1=NC2(N=C1N)c1cc(Br)ccc1CCC21CC1},   avoiding pyrophoric materials (substances that ignite in moisture or air) in   your synthesis route.                     \\
S34  &  \mbox{P}     & Plan a synthesis route for   \texttt{COc1cc2ncc3c(N)nc(-c4cncc\allowbreak{}(OCCN(Cc5ccc(F)cc5)C(=O)OC(C)(C)C)c4)cc3c2cc1OC}   that uses no pyrophoric or water-reactive substances.                                 \\
S35  &  \mbox{P}     & Perform synthesis planning for   \texttt{O{[}C@H{]}1C{[}C@H{]}\allowbreak{}(c2cnn3c\allowbreak{}(N{[}C@H{]}4CCc5ccccc54)ncnc23)\allowbreak{}C=C1COCc1ccccc1} without using pyrophoric substances in your synthesis plan.                     \\
S36  &  \mbox{P}     & Synthesize \texttt{CC(=O)N1c2ccc(N3CCNCC3)\allowbreak{}cc2{[}C@H{]}(Nc2ccccc2)\allowbreak{}{[}C@@H{]}(C)\allowbreak{}{[}C@@H{]}1C}   without using any pyrophoric or water-reactive reagents.                                                 \\
S37  &  \mbox{P}     & Find the shortest synthesis route for   \texttt{COc1cc2c(=O){[}nH{]}c(=O)n\allowbreak{}({[}C@@H{]}3O{[}C@H{]}(CO){[}C@H{]}4OC(C)(C)O{[}C@H{]}43)c2cc1OC},   avoiding pyrophoric substances.                         \\
S38 &  \mbox{P}     & Identify a synthesis route for   \texttt{Oc1ccc2c3c(ccc2c1)\allowbreak{}Cc1ccccc1OC3c1ccc\allowbreak{}(OCCN2CCCCC2)cc1} that does not use   pyrophoric substances.                                                               \\
S39 &  \mbox{P}     & Plan the synthesis of   \texttt{C{[}C@@H{]}(O)C{[}C@H{]}1OC{[}C@@H{]}\allowbreak{}(C2CCCCC2)N(c2cc(C\#CC(C)(C)C)sc2C(=O)O)C1=O}. Do   not use any pyrophoric substances.                                            \\
S40 &  \mbox{P}     & Find the shortest synthesis route for   \texttt{OC{[}C@H{]}1C{[}C@@H{]}\allowbreak{}(c2cnn3c\allowbreak{}(N{[}C@H{]}4CCc5ccccc54)\allowbreak{}ncnc23)C{[}C@@H{]}1O} that avoids   using any pyrophoric substances.                            \\
\midrule
S41  &  \mbox{U} & Find the shortest synthesis path for \texttt{C{[}C@@H{]}1CCCN1CCc1nnc2cc(Br)ccc2c1O},   but avoid using \texttt{C=C{[}Sn{]}(CCCC)(CCCC)CCCC} in your synthesis route.                                         \\
S42  &  \mbox{U} & Plan a synthesis route for   \texttt{CC(=O)NC{[}C@H{]}1CN(c2ccc3c(c2)\allowbreak{}CCCc2c(C(C)C)\allowbreak{}n{[}nH{]}c2-3)C(=O)O1}. Avoid using   phosgene (\texttt{O=C(Cl)Cl}) in your synthesis.                                      \\
S43  &  \mbox{U} & Find the shortest synthesis route for   \texttt{C{[}C@@H{]}1CNC\allowbreak{}(=O)c2cc3cc\allowbreak{}(OCCCN4CCCCC4)ccc3n21} that does not use trimethyl   borate (\texttt{COB(OC)OC}).                                                   \\
S44  &  \mbox{U} & Plan the synthesis of   \texttt{COCCCc1cc(CN(C(=O){[}C@H{]}2CNCC{[}C@@H{]}2c2ccc\allowbreak{}(OCCOc3c(Cl)cc(C)cc3Cl)cc2)C2CC2)cc(OCCOC)c1}   without using hexane (\texttt{CCCCCC}).                                       \\
S45  &  \mbox{U} & Identify a synthesis plan for   \texttt{COc1cc2c\allowbreak{}(=O){[}nH{]}c(=O)n({[}C@@H{]}3O{[}C@H{]}\allowbreak{}(CO){[}C@H{]}4OC(C)\allowbreak{}(C)O{[}C@H{]}43)\allowbreak{}c2cc1OC} that   does not use methanol (\texttt{CO}).                               \\
S46  &  \mbox{U} & Plan the synthesis of   \texttt{CC(C)c1ccc2c\allowbreak{}(c1)OC1(O)c3ccccc3C(=O)\allowbreak{}C21NC(=O)c1cc(-c2ccccc2)n{[}nH{]}1}. Do not use phenol (\texttt{Oc1ccccc1}).                                                             \\
S47  &  \mbox{U} & Synthesize \texttt{CCOC(=O)/C(N)=N/Nc1cc(Cl)ccc1{[}N+{]}(=O){[}O-{]}} without using   nitric acid (\texttt{O={[}N+{]}({[}O-{]})O}).                                                                           \\
S48  &  \mbox{U} & Find the shortest synthesis route for   \texttt{C\#CC1(O)C(C)=CC2(CC1(C)C(F)(F)F)OC(C)C(C)O2} that does not use toluene   (\texttt{Cc1ccccc1}).                                                               \\ 
\midrule
M1 & \mbox{C \& U} & Find a synthesis route for \texttt{C4=C(C(C2C(C1=CC=NC=C1)C2)(C3=CC=CC=C3)O)C=CC=C4} that avoids all known or suspected carcinogens and \texttt{Cl}. 
\\ 
M2 & \mbox{C \& U} & Find the shortest synthesis path for \texttt{Cn1oc(=O)nc1/C(=N\textbackslash OCc1cccc(N)n1)c1ccccc1} that does not use carcinogenic substances and \texttt{NO}. 
\\
M3 & \mbox{C \& U} & Generate a synthesis plan without carcinogens for \texttt{C2=C(C1=C(C=CC=C1)[N]2N(CCN(C)C)C)C3=CC=CC=C3}. Also avoid \texttt{BrBr}. \\ 
M4 & \mbox{C \& P \& U} & Generate a synthesis plan without carcinogens or pyrophoric substances for \texttt{C2=C(C(C1=CC=CC=C1)(CCN(C)C)C(CC)=O)C=CC=C2}. Also avoid \texttt{Br}. \\ 
M5 & \mbox{C \& P \& U} & Find the shortest synthesis path for \texttt{C1=C(O)C=CC4=C1C35C(C(N(CCC2=CC=CC=C2)CC3)C4)CCCC5}. Do not use any carcinogens, pyrophoric materials, or \texttt{Cl}. \\ 
M6 & \mbox{P \& U} & Plan the shortest synthesis plan for \texttt{C1=C(Br)SC2=C1C(=NCC3=NN=C([N]23)C)C4=C(C=CC=C4)Cl} that avoids all pyrophoric and water-reactive substances and \texttt{BrBr}. \\ 
M7 & \mbox{C \& P \& U} & Plan the synthesis of \texttt{C2=C(C(C1=CC=CC=C1)(C(OC(=O)C)CC)CC(N(C)C)C)C=CC=C2} but avoid using carcinogen material, pyrophoric material and \texttt{Br}. \\ 
M8 & \mbox{C \& U} & Design a synthesis path for \texttt{OCCN1CCC(c2ccc3c(c2)-n2nc(-c4ncnn4CC(F)(F)F)cc2CCO3)CC1}. Do not use any carcinogens or \texttt{BrBr}. \\ 
M9 & \mbox{C \& U} & Find a synthesis route for \texttt{O.N[C@@H](C(=O)NC1C2CCC(=C(N2C1=O)C(O)=O)Cl)c3ccccc3} that avoids all known or suspected carcinogens and \texttt{Br}. \\ 
M10 & \mbox{C \& U} & Find a synthesis route for \texttt{C1=C4C(=C2C(=C1)C=CC=C2)CC3CN(C)CC(C3(O4)O)C} that avoids all known or suspected carcinogens and \texttt{Br}. \\ 
M11 & \mbox{P \& U} & Synthesize \texttt{CCOC(OCC)C(=O)OCC(=O)C1(O)CC(OC2CC(N)\allowbreak{}C(O)C(C)O2)c3c(O)c4C(=O)c5c(OC)cccc5C(=O)c4c(O)c3C1} without using any pyrophoric or water-reactive reagents and \texttt{Br} and \texttt{CO}. \\ 
M12 & \mbox{C \& P} & Plan the shortest synthesis route for \texttt{CCN(CC)C(C)=NN=Cc1c(O)c2c3C(=O)C4(C)OC=CC(OC)C(C)C(OC(C)=O)\allowbreak{}C(C)C(O)C(C)C(O)C(C)C=CC=C(C)C(=O)Nc1c(O)c2c(O)c(C)c3O4} without using any carcinogens or pyrophoric substances. \\ 
M13 & \mbox{P \& U} & Find the shortest synthesis route for \texttt{O.C[C@H](O)[C@@H]1[C@H]2CC(=C(N2C1=O)C(O)=O)SCCN=CN} that avoids using any pyrophoric substances and \texttt{Cl}. \\ 
M14 & \mbox{C \& U} & Find the synthesis path for \texttt{C1=C(O)C=CC4=C1C3(C(C(N(CC2CC2)CC3)C4)(C)C)CC} but avoid using carcinogen material and \texttt{BrBr} and \texttt{CO}. \\ 
M15 & \mbox{C \& U} & Plan the shortest synthesis route for \texttt{N\#Cc5ccc4Oc1ccccc1C2=C(CCN(CC2)CC3CCCC3)c4c5} without using carcinogenic materials or \texttt{BrBr}. \\ 
M16 & \mbox{C \& U} & Find a synthesis route for \texttt{COc1cc2ncc3c(N)nc(-c4cncc(OCCNCc5ccc(F)cc5)c4)cc3c2cc1OC} that avoids all known or suspected carcinogens and \texttt{BrBr}. \\ 
M17 & \mbox{C \& U} & Find the synthesis path for \texttt{COc1cccc2C(=O)c3c(O)c4CC(O)(CC(O)c4c(O)c3C(=O)c12)C(=O)CO} but avoid using carcinogen material and \texttt{CO}. \\ 
M18 & \mbox{C \& U} & Find the synthesis path for \texttt{O=C(CO)N1CCC(c2ccc3c(c2)-n2nc(-c4ncnn4CC(F)(F)F)cc2CCO3)CC1} but avoid using carcinogen material and \texttt{BrBr}. \\ 
M19 & \mbox{C \& P \& U} & Find the shortest synthesis path for \texttt{COc1ccc2C[C@H]3[C@H]4CCCC[C@@]4(CCN3C)c2c1}. Do not use any carcinogens, pyrophoric materials, or \texttt{BrBr} and \texttt{CO}. \\ 
M20 & \mbox{C \& U} & Generate a synthesis plan without carcinogens for \texttt{[Br-].CC(C)[N+](C)(CCOC(=O)C1c2ccccc2Oc3ccccc13)C(C)C}. Also avoid \texttt{Br}. \\ 
M21 & \mbox{C \& U} & Plan the synthesis of \texttt{c1ccc2Oc3c(cc(cc3)Cl)[C@@H]3[C@@H](c2c1)C[N@](CC3)C}. Avoid carcinogenic materials and \texttt{Oc1ccccc1}. \\ 
M22 & \mbox{C \& U} & Plan the synthesis of \texttt{CN1CCN(CC/C=C/2c3ccccc3Sc4ccc(cc24)[S](=O)(=O)N(C)C)CC1}. Avoid carcinogenic materials and \texttt{Cl} and \texttt{O=C1CCC(=O)N1Br}. \\ 
M23 & \mbox{C \& U} & Generate a synthesis plan without carcinogens for \texttt{Oc5ccc4CC2C1CCOCC1(CCN2CC3CC3)c4c5}. Also avoid \texttt{CO}. \\ 
M24 & \mbox{C \& U} & Plan the shortest synthesis route for \texttt{C1=C(C\#N)C=CC3=C1C4=C(C2=C(C=CC=C2)S3)CCN(CC4)C} without using carcinogenic materials or \texttt{O=C1CCC(=O)N1Br}. \\ 
M25 & \mbox{C \& P \& U} & Find the synthesis path for \texttt{C1=C(SC2=C1C(=NCC3=NN=C([N]23)C4CCCCC4)C5=CC=CC=C5Cl)Br} but avoid using carcinogen material, pyrophoric material and \texttt{BrBr}. \\ 
M26 & \mbox{C \& P \& U} & Plan the shortest possible synthesis route for \texttt{C1=C(O)C=CC3=C1C24C(C(NCC2)C3)CCCC4}, but do not use any carcinogens, pyrophoric substances, or \texttt{Cl} in the route. \\ 
M27 & \mbox{C \& U} & Plan the shortest synthesis route for \texttt{CCCN(CCC)CCc1ccc(c2c1CC(N2)=C)O} without using carcinogenic materials or \texttt{BrBr} and \texttt{O=C1CCC(=O)N1Br}. \\ 
M28 & \mbox{C \& U} & Find the shortest synthesis path for \texttt{c1(ccc(c(c1)Cl)Cl)CC(N1[C@H](C[C@]2(CC1)NC(NC2=O)=O)\allowbreak{}CN1CCCC1)=O} that does not use carcinogenic substances and \texttt{Br}. \\ 
M29 & \mbox{C \& P \& U} & Find the synthesis path for \texttt{COC[C@H](C)COCc1ccc([C@@]2(O)CCN(C(=O)OC(C)\allowbreak{}(C)C)C[C@@H]2c2noc(-c3ccccc3CCNC(C)=O)c2Br)cc1} but avoid using carcinogen material, pyrophoric material and \texttt{BrBr} and \texttt{NO}. \\ 
M30 & \mbox{C \& P} & Identify a synthesis route for \texttt{C1=C(OC)C(=CC2=C1C(=NN=C(C2)C)C3=CC=CC(=C3)Cl)OC} that does not use carcinogenic or pyrophoric materials. \\ 
M31 & \mbox{C \& U} & Find the shortest synthesis path for \texttt{[C@]2(C1=CC(=CC=C1)O)([C@H](CN(C)CC2)C)CCC} that does not use carcinogenic substances and \texttt{Br}. \\ 
M32 & \mbox{P \& U} & Identify a synthesis route for \texttt{C[C@H](c1ccccc1)N1C[C@H]2CC=C[C@@]2(C(=O)OC(C)(C)C)C1} that does not use pyrophoric substances or \texttt{C=CCCl} and \texttt{CO}. \\ 
M33 & \mbox{C \& U} & Find the shortest synthesis path for \texttt{CN1CC[C@]23CCCC[C@H]2[C@H]1Cc4ccc(O)cc34} that does not use carcinogenic substances and \texttt{Oc1ccccc1}. \\ 
M34 & \mbox{P \& U} & Synthesize \texttt{COC1=C(N3C(SC1)C(NC(=O)C(N)C2C=CCC=C2)C3=O)C(O)=O} without using any pyrophoric or water-reactive reagents and \texttt{CO}. \\ 
M35 & \mbox{C \& U} & Generate a synthesis plan without carcinogens for \texttt{COC[C@H](C)COCc1ccc([C@@]2(O)CCN(C(=O)OC(C)(C)C)\allowbreak{}C[C@@H]2CO)cc1}. Also avoid \texttt{Cl} and \texttt{O=C1CCC(=O)N1Br}. \\ 
M36 & \mbox{P \& U} & Find the shortest synthesis route for \texttt{Oc1ccc2c(c1)[C@H](c1ccc(OCCN3CCCC3)cc1)[C@H](c1ccccc1)CO2} that avoids using any pyrophoric substances and \texttt{CO}, \texttt{O=C1CCC(=O)N1Br}, \texttt{Oc1ccc(O)cc1} and \texttt{Oc1ccccc1}. \\ 
M37 & \mbox{C \& U} & Plan the shortest synthesis route for \texttt{C1=CC(=C3C2=C1CC5C4C2(C(O3)C(O)C=C4)CCN5)OC} without using carcinogenic materials or \texttt{BrBr}, \texttt{CO} and \texttt{O=C1CCC(=O)N1Br}. \\ 
M38 & \mbox{C \& U} & Plan the synthesis of \texttt{[C@H]13[C@H](C[C@@H]1C2=CC=C(F)C=C2)CN(C3)CCN4C(=O)C5=C\allowbreak{}(NC4=O)C=CC=C5}. Avoid carcinogenic materials and \texttt{Cl}. \\ 
M39 & \mbox{C \& U} & Find a synthesis route for \texttt{[C@H](O)(/C=C/C1=C(C3=C(OC12CCCC2)C=CC=C3)C4=CC=C(F)C=C4)\allowbreak{}C[C@H](O)CC(OCC)=O} that avoids all known or suspected carcinogens and \texttt{CO} and \texttt{O=C1CCC(=O)N1Br}. \\ 
M40 & \mbox{C \& U} & Design a synthesis path for \texttt{COc1cccc2C(=O)c3c(O)\allowbreak{}c4C[C@@](O)(C[C@@H](OC5CC(N)CC(C)O5)c4c(O)c3C(=O)c12)C(=O)CO}. Do not use any carcinogens or \texttt{CO}. \\ 
M41 & \mbox{C \& U} & Generate a synthesis plan without carcinogens for \texttt{[C@H]15C3=C(CCC2=C1C=CC=C2)C=CC=C3[C@H]4C[C@](C(C)C)(O)CCN4C5}. Also avoid \texttt{Br} and \texttt{CO}. \\ 
M42 & \mbox{C \& U} & Find the synthesis path for \texttt{C1=CC(=C3C2=C1CC5C4C2(C(O3)C(O)C=C4)CCN5)O} but avoid using carcinogen material and \texttt{Br}, \texttt{BrBr} and \texttt{O=C1CCC(=O)N1Br}. \\ 
M43 & \mbox{C \& U} & Find the shortest synthesis path for \texttt{COC[C@H](C)COCc1ccc([C@@]2(O)CCN(C(=O)OC(C)(C)C)\allowbreak{}C[C@@H]2C=O)cc1} that does not use carcinogenic substances and \texttt{O=C1CCC(=O)N1Br}. \\ 
M44 & \mbox{C \& U} & Plan the synthesis of \texttt{C[C@H]1[C@H](NC(=O)C(=N/OC(C)(C)C(O)=O)\textbackslash c2csc([NH3+])n2)\allowbreak{}C(=O)N1[S]([O-])(=O)=O}. Avoid carcinogenic materials and \texttt{BrBr} and \texttt{CO}. \\ 
M45 & \mbox{C \& U} & Plan the synthesis of \texttt{C1=C(O)C=CC4=C1C3(C(C(N(CC2CCC2)CC3)C4)(C)C)CC}. Avoid carcinogenic materials and \texttt{BrBr}. \\ 
M46 & \mbox{C \& P \& U} & Design a synthesis path for \texttt{COC[C@H](C)COCc1ccc([C@@]2(O)CCNC[C@@H]2c2noc(-c3ccccc3CCNC\allowbreak{}(C)=O)\allowbreak{}c2Br)cc1}. Do not use any carcinogens, pyrophoric substances, or \texttt{BrBr}. \\ 
M47 & \mbox{C \& U} & Find a synthesis route for \texttt{CN(C)Cc1oc(cc1)CSCCNC=1NC=C(CN1)Cc1cnc(cc1)C} that avoids all known or suspected carcinogens and \texttt{O=C1CCC(=O)N1Br}. \\ 
M48 & \mbox{C \& U} & Find the shortest synthesis path for \texttt{[C@H]4(CC3C2=C1C(=C[N](C1=CC=C2)C(C)C)CC3N(C4)C)C\allowbreak{}(NC5CCCCC5)=O} that does not use carcinogenic substances and \texttt{Br}. \\ 
M49 & \mbox{C \& U} & Design a synthesis path for \texttt{C1=C2C(=CC=C1)CCC(=C2)CC3=NCNC3}. Do not use any carcinogens or \texttt{Br}. \\ 
M50 & \mbox{P \& U} & Identify a synthesis route for \texttt{[Na+].Cn1nnnc1SCC2=C(N3[C@H](SC2)[C@H](NC(=O)CSC(F)(F)F)\allowbreak{}C3=O)C([O-])=O} that does not use pyrophoric substances or \texttt{Br}. \\ 
M51 & \mbox{C \& U} & Design a synthesis path for \texttt{C1=C2C(=CC=C1)N(C(Cl)(CN=C2C3=CC=CC=C3Cl)CO)C}. Do not use any carcinogens or \texttt{CO}. \\ 
M52 & \mbox{C \& U} & Plan the shortest synthesis route for \texttt{[C@@H]3(N(C1CCC1)C2CCC2)CC4=C(OC3)C(=CC=C4C(=O)N)F} without using carcinogenic materials or \texttt{BrBr}. \\ 
M53 & \mbox{C \& U} & Find a synthesis route for \texttt{CC(C)(C)OC(=O)NC1(C2CCc3cc(Sc4cccc(OCc5ccccc5)c4)ccc3C2)\allowbreak{}COC(C)(C)OC1} that avoids all known or suspected carcinogens and \texttt{CO}. \\ 
M54 & \mbox{C \& U} & Find the shortest synthesis path for \texttt{Cn1oc(=O)nc1/C(=N\textbackslash OCc1cccc(NC(=O)OCCc2ccccc2)n1)c1ccccc1} that does not use carcinogenic substances and \texttt{NO}. \\ 
M55 & \mbox{C \& U} & Plan the synthesis of \texttt{CS(=O)(=O)CCN1CCC(c2ccc3c(c2)-n2nc(-c4ncnn4CC(F)(F)F)\allowbreak{}cc2CCO3)CC1}. Avoid carcinogenic materials and \texttt{BrBr}. \\ 
M56 & \mbox{C \& U} & Find a synthesis route for \texttt{NC(=O)CN1CCC(c2ccc3c(c2)-n2nc(-c4ncnn4CC(F)(F)F)cc2CCO3)CC1} that avoids all known or suspected carcinogens and \texttt{BrBr}. \\ 
M57 & \mbox{C \& U} & Plan the shortest synthesis route for \texttt{C4=C(C3=C(CCN2CCN(C1=C(C=CC=C1)OC)CC2)OC(N3)=O)C=CC(=C4)F} without using carcinogenic materials or \texttt{O=C1CCC(=O)N1Br}. \\ 
M58 & \mbox{C \& U} & Find the synthesis path for \texttt{[C@]134[C@@H]([C@H](CC2=C1C=C(O)C=C2)N(C)CC3)CCCC4} but avoid using carcinogen material and \texttt{Oc1ccccc1}. \\ 
M59 & \mbox{C \& U} & Find a synthesis route for \texttt{[C@]134[C@@H]([C@H](CC2=C1C=C(C)C=C2)N(C)CC3)CCCC4} that avoids all known or suspected carcinogens and \texttt{Cc1ccccc1}. \\ 
M60 & \mbox{C \& U} & Find a synthesis route for \texttt{C=C(C[C@@H](Cc1ccc(-c2ccccc2)cc1)NC(=O)OC(C)(C)C)C(=O)OC} that avoids all known or suspected carcinogens and \texttt{Br} and \texttt{CO}. \\ 
M61 & \mbox{C \& U} & Design a synthesis path for \texttt{C1=CC=CC3=C1C(C2=CC=CC=C2)(C=C3)CCN(C)C}. Do not use any carcinogens or \texttt{Br}. \\ 
M62 & \mbox{C \& P} & Design a synthesis path for \texttt{C2=C1C(=NN=C(CC1=CC(=C2OC)OC)C)C3=CC=C(N)C=C3}. Do not use any carcinogens or pyrophoric substances. \\ 
M63 & \mbox{C \& U} & Generate a synthesis plan without carcinogens for \texttt{C4=C(C(C3CCN(CCC1=C(N=C2N(C1=O)CCS2)C)CC3)=O)C=CC(=C4)F}. Also avoid \texttt{Cl}. \\
\bottomrule
\end{longtable}
}

\subsection{\gmo Tasks}

{Table~\ref{tab:dataset:general} shows the products for the \gmo tasks in the benchmark dataset.  There is one common instruction for all of these tasks, shown here:}

\begin{promptbox}
Plan the best possible route considering all available safety and cost metrics.
\end{promptbox}

{
\renewcommand{\arraystretch}{1.2}
\begin{longtable}{>{\centering\arraybackslash}p{0.05\linewidth}
>{\raggedright\arraybackslash}p{0.85\linewidth}}
\caption{\gmo Final Products}
\label{tab:dataset:general}\\
\toprule
ID & Target Molecule (Product) \\
\midrule
\endfirsthead

\toprule
ID & Target Molecule (Product) \\
\midrule
\endhead

\midrule
\multicolumn{2}{r}{\emph{Continued on next page}}\\
\midrule
\endfoot

\hline
\endlastfoot

1 & \texttt{C\#CC1(O)C(C)=CC2(CC1(C)C(F)(F)F)OC(C)C(C)O2} \\
2 & \texttt{C1=C(Br)SC2=C1C(=NCC3=NN=C([N]23)C)C4=C(C=CC=C4)Cl} \\
3 & \texttt{C1=C(C\#N)C=CC3=C1C4=C(C2=C(C=CC=C2)S3)CCN(CC4)C} \\
4 & \texttt{C1=C(O)C=CC3=C1C24C(C(NCC2)C3)CCCC4} \\
5 & \texttt{C1=C(O)C=CC4=C1C3(C(C(N(CC2CC2)CC3)C4)(C)C)CC} \\
6 & \texttt{C1=C(O)C=CC4=C1C3(C(C(N(CC2CCC2)CC3)C4)(C)C)CC} \\
7 & \texttt{C1=C(O)C=CC4=C1C35C(C(N(CCC2=CC=CC=C2)CC3)C4)CCCC5} \\
8 & \texttt{C1=C(OC)C(=CC2=C1C(=NN=C(C2)C)C3=CC=CC(=C3)Cl)OC} \\
9 & \texttt{C1=C(SC2=C1C(=NCC3=NN=C([N]23)C4CCCCC4)C5=CC=CC=C5Cl)Br} \\
10 & \texttt{C1=C2C(=CC=C1)CCC(=C2)CC3=NCNC3} \\
11 & \texttt{C1=C2C(=CC=C1)N(C(Cl)(CN=C2C3=CC=CC=C3Cl)CO)C} \\
12 & \texttt{C1=C4C(=C2C(=C1)C=CC=C2)CC3CN(C)CC(C3(O4)O)C} \\
13 & \texttt{C1=CC(=C3C2=C1CC5C4C2(C(O3)C(O)C=C4)CCN5)O} \\
14 & \texttt{C1=CC(=C3C2=C1CC5C4C2(C(O3)C(O)C=C4)CCN5)OC} \\
15 & \texttt{C1=CC=CC3=C1C(C2=CC=CC=C2)(C=C3)CCN(C)C} \\
16 & \texttt{C2=C(C(C1=CC=CC=C1)(C(OC(=O)C)CC)CC(N(C)C)C)C=CC=C2} \\
17 & \texttt{C2=C(C(C1=CC=CC=C1)(CCN(C)C)C(CC)=O)C=CC=C2} \\
18 & \texttt{C2=C(C1=C(C=CC=C1)[N]2N(CCN(C)C)C)C3=CC=CC=C3} \\
19 & \texttt{C2=C1C(=NN=C(CC1=CC(=C2OC)OC)C)C3=CC=C(N)C=C3} \\
20 & \texttt{C4=C(C(C2C(C1=CC=NC=C1)C2)(C3=CC=CC=C3)O)C=CC=C4} \\
21 & \texttt{C4=C(C(C3CCN(CCC1=C(N=C2N(C1=O)CCS2)C)CC3)=O)C=CC(=C4)F} \\
22 & \texttt{C4=C(C3=C(CCN2CCN(C1=C(C=CC=C1)OC)CC2)OC(N3)=O)C=CC(=C4)F} \\
23 & \texttt{C=C(C[C@@H](Cc1ccc(-c2ccccc2)cc1)NC(=O)OC(C)(C)C)C(=O)O} \\
24 & \texttt{C=C(C[C@@H](Cc1ccc(-c2ccccc2)cc1)NC(=O)OC(C)(C)C)C(=O)OC} \\
25 & \texttt{CC\#CCn1c(Br)nc(C=O)c1C(=O)OC} \\
26 & \texttt{CC(=O)N1c2ccc(N3CCNCC3)cc2[C@H](Nc2ccccc2)[C@@H](C)[C@@H]1C} \\
27 & \texttt{CC(=O)NC[C@H]1CN(c2ccc3c(c2)CCCc2c(C(C)C)n[nH]c2-3)C(=O)O1} \\
28 & \texttt{CC(=O)OCc1nc2cnc(Br)cc2n1C(C)(C)COC(C)=O} \\
29 & \texttt{CC(=O)c1ccc2c(c1)C=CC(O)(CO)CO2} \\
30 & \texttt{CC(C)(C)OC(=O)N1CC=C(c2ccc3c(c2)-n2nc(-c4ncnn4CC(F)(F)F)cc2CCO3)CC1} \\
31 & \texttt{CC(C)(C)OC(=O)NC1(C2CCc3cc(Sc4cccc(OCc5ccccc5)c4)ccc3C2)COC(C)(C)OC1} \\
32 & \texttt{CC(C)(C)OC(=O)NC1(c2nc(NCc3ccccn3)c3c(Cl)ccn3n2)CC1} \\
33 & \texttt{CC(C)(C)OC(=O)NC1(c2nc(O)c3c(Cl)ccn3n2)CC1} \\
34 & \texttt{CC(C)(C)OC(=O)N[C@@H]1c2cccnc2C(=O)CC[C@H]1c1cccc(F)c1F} \\
35 & \texttt{CC(C)(C)OC(=O)N[C@@H]1c2cccnc2[C@H](N)CC[C@H]1c1cccc(F)c1F} \\
36 & \texttt{CC(C)(C)OC(=O)N[C@@H]1c2cccnc2[C@H](O)CC[C@H]1c1cccc(F)c1F} \\
37 & \texttt{CC(C)(CO)n1c(CO)nc2cnc(Br)cc21} \\
38 & \texttt{CC(C)c1ccc2c(c1)OC1(O)c3ccccc3C(=O)C21NC(=O)c1cc(-c2ccccc2)n[nH]1} \\
39 & \texttt{CC1=NC2(N=C1N)c1cc(Br)ccc1CCC21CC1} \\
40 & \texttt{CCCC[Sn](/C=C/C1(O)C(C)=CC2(CC1(C)C(F)(F)F)OC(C)C(C)O2)(CCCC)CCCC} \\
41 & \texttt{CCCN(CCC)CCc1ccc(c2c1CC(N2)=C)O} \\
42 & \texttt{{CCN(CC)C(C)=NN=Cc1c(O)c2c3C(=O)C4(C)OC=CC(OC)C(C)\allowbreak{}C(OC(C)=O)C(C)C(O)C(C)C(O)C(C)C=CC=C(C)C(=O)Nc1c(O)c2c(O)c(C)c3O4}} \\
43 & \texttt{CCOC(=O)/C(N)=N/Nc1cc(Cl)ccc1[N+](=O)[O-]} \\
44 & \texttt{{CCOC(OCC)C(=O)OCC(=O)C1(O)CC(OC2CC(N)C(O)C(C)O2)c3c\allowbreak{}(O)c4C(=O)c5c(OC)cccc5C(=O)c4c(O)c3C1}} \\
45 & \texttt{CC[C@@H](OC(=O)c1ccccc1)[C@H]1CCCN(C(=O)OC(C)(C)C)C1} \\
46 & \texttt{CN(C)Cc1oc(cc1)CSCCNC=1NC=C(CN1)Cc1cnc(cc1)C} \\
47 & \texttt{CN1CCN(CC/C=C/2c3ccccc3Sc4ccc(cc24)[S](=O)(=O)N(C)C)CC1} \\
48 & \texttt{CN1CC[C@]23CCCC[C@H]2[C@H]1Cc4ccc(O)cc34} \\
49 & \texttt{COC(=O)CCc1cc2cc(-c3noc(-c4ccc(OC(C)C)c(Cl)c4)n3)ccc2n1C} \\
50 & \texttt{COC(=O)c1ccc2c(c1)C=CC(=C(Cl)Cl)CO2} \\
51 & \texttt{COC(=O)c1ccc2c(c1)C=CC(=CCl)CO2} \\
52 & \texttt{COC1=C(N3C(SC1)C(NC(=O)C(N)C2C=CCC=C2)C3=O)C(O)=O} \\
53 & \texttt{\makecell[l]{COCCCc1cc(CN(C(=O)[C@H]2CN(C(=O)OC(C)(C)C)CC[C@@H]2c2ccc(OCCOc3c\\(Cl)cc(C)cc3Cl)cc2)C2CC2)cc(OCCOC)c1}} \\
54 & \texttt{\makecell[l]{COCCCc1cc(CN(C(=O)[C@H]2CNCC[C@@H]2c2ccc(OCCOc3c(Cl)cc(C)cc3Cl)\\cc2)C2CC2)cc(OCCOC)c1}} \\
55 & \texttt{COC[C@H](C)COCc1ccc([C@@]2(O)CCN(C(=O)OC(C)(C)C)C[C@@H]2C=O)cc1} \\
56 & \texttt{COC[C@H](C)COCc1ccc([C@@]2(O)CCN(C(=O)OC(C)(C)C)C[C@@H]2CO)cc1} \\
57 & \texttt{\makecell[l]{COC[C@H](C)COCc1ccc([C@@]2(O)CCN(C(=O)OC(C)(C)C)C[C@@H]2c2noc\\(-c3ccccc3CCNC(C)=O)c2Br)cc1}} \\
58 & \texttt{COC[C@H](C)COCc1ccc([C@@]2(O)CCNC[C@@H]2c2noc(-c3ccccc3CCNC(C)=O)\allowbreak{}c2Br)cc1} \\
59 & \texttt{COc1cc2c(=O)[nH]c(=O)n([C@@H]3O[C@H](CO)[C@H]4OC(C)(C)O[C@H]43)\allowbreak{}c2cc1OC} \\
60 & \texttt{COc1cc2c(Oc3cc(C)c(C)nc3-c3cccc(C)n3)ccnc2cc1OCCNCCO} \\
61 & \texttt{COc1cc2ncc3c(N)nc(-c4cncc(OCCN(Cc5ccc(F)cc5)C(=O)OC(C)(C)C)c4)\allowbreak{}cc3c2cc1OC} \\
62 & \texttt{COc1cc2ncc3c(N)nc(-c4cncc(OCCNCc5ccc(F)cc5)c4)cc3c2cc1OC} \\
63 & \texttt{COc1ccc2C[C@H]3[C@H]4CCCC[C@@]4(CCN3C)c2c1} \\
64 & \texttt{COc1cccc2C(=O)c3c(O)c4CC(O)(CC(O)c4c(O)c3C(=O)c12)C(=O)CO} \\
65 & \texttt{\makecell[l]{COc1cccc2C(=O)c3c(O)c4C[C@@](O)(C[C@@H](OC5CC(N)CC\allowbreak{}(C)O5)c4c(O)c3C(=O)\\c12)C(=O)CO}} \\
66 & \texttt{CS(=O)(=O)CCN1CCC(c2ccc3c(c2)-n2nc(-c4ncnn4CC(F)(F)F)cc2CCO3)CC1} \\
67 & \texttt{C[C@@H](O)C[C@H]1OC[C@@H](C2CCCCC2)N(c2cc(C\#CC(C)(C)C)sc2C(=O)O)C1=O} \\
68 & \texttt{C[C@@H](O)c1nc2cnc3ccsc3c2n1[C@H]1CC[C@H](CO)CC1} \\
69 & \texttt{C[C@@H]1CCCN1CCc1nnc2cc(Br)ccc2c1O} \\
70 & \texttt{C[C@@H]1CNC(=O)c2cc3cc(OCCCN4CCCCC4)ccc3n21} \\
71 & \texttt{C[C@H](O[Si](C)(C)C(C)(C)C)[C@@H]1CC(=O)CC(C)(C)N1} \\
72 & \texttt{C[C@H](c1ccccc1)N1C[C@H]2CC=C[C@@]2(C(=O)OC(C)(C)C)C1} \\
73 & \texttt{C[C@H](c1ccccc1)N1C[C@]2(C(=O)OC(C)(C)C)C=CC[C@@H]2C1=S} \\
74 & \texttt{\makecell[l]{C[C@H]1[C@H](NC(=O)C(=N/OC(C)(C)C(O)=O)\textbackslash{}c2csc([NH3+])n2)C(=O)N1[S]\\([O-])(=O)=O}} \\
75 & \texttt{C[Si](C)(C)CCOCn1cc(C2CCc3c(C(=O)O)nn(COCC[Si](C)(C)C)c3C2)cn1} \\
76 & \texttt{ClCc1ccc2c(c1)Nc1nccnc1S2} \\
77 & \texttt{Cn1oc(=O)nc1/C(=N\textbackslash{}OCc1cccc(N)n1)c1ccccc1} \\
78 & \texttt{Cn1oc(=O)nc1/C(=N\textbackslash{}OCc1cccc(NC(=O)OCCc2ccccc2)n1)c1ccccc1} \\
79 & \texttt{Cn1oc(=O)nc1/C(=N\textbackslash{}\textbackslash{}OCc1cccc(N)n1)c1ccccc1} \\
80 & \texttt{Cn1oc(=O)nc1/C(=N\textbackslash{}\textbackslash{}OCc1cccc(NC(=O)OCCc2ccccc2)n1)c1ccccc1} \\
81 & \texttt{FC(F)(F)Cn1ncnc1-c1cc2n(n1)-c1cc(C3CCNCC3)ccc1OCC2} \\
82 & \texttt{N\#Cc5ccc4Oc1ccccc1C2=C(CCN(CC2)CC3CCCC3)c4c5} \\
83 & \texttt{NC(=O)CN1CCC(c2ccc3c(c2)-n2nc(-c4ncnn4CC(F)(F)F)cc2CCO3)CC1} \\
84 & \texttt{O.C[C@H](O)[C@@H]1[C@H]2CC(=C(N2C1=O)C(O)=O)SCCN=CN} \\
85 & \texttt{O.N[C@@H](C(=O)NC1C2CCC(=C(N2C1=O)C(O)=O)Cl)c3ccccc3} \\
86 & \texttt{O=C(CO)N1CCC(c2ccc3c(c2)-n2nc(-c4ncnn4CC(F)(F)F)cc2CCO3)CC1} \\
87 & \texttt{O=C(Nc1cccc(Cl)c1)N1CCc2[nH]nc(C(=O)N3CC(F)CO3)c2C1} \\
88 & \texttt{O=C(OCc1ccccc1)N1CC[C@H]2CCCN(CCc3ccccc3)C[C@H]21} \\
89 & \texttt{O=S(=O)(C\#Cc1ccc(Cl)cc1)N1CCNCC1} \\
90 & \texttt{OCCN1CCC(c2ccc3c(c2)-n2nc(-c4ncnn4CC(F)(F)F)cc2CCO3)CC1} \\
91 & \texttt{OC[C@H]1C[C@@H](c2cnn3c(N[C@H]4CCc5ccccc54)ncnc23)C[C@@H]1O} \\
92 & \texttt{O[C@H]1C[C@H](c2cnn3c(N[C@H]4CCc5ccccc54)ncnc23)C=C1COCc1ccccc1} \\
93 & \texttt{Oc1ccc2c(c1)[C@H](c1ccc(OCCN3CCCC3)cc1)[C@H](c1ccccc1)CO2} \\
94 & \texttt{Oc1ccc2c3c(ccc2c1)Cc1ccccc1OC3c1ccc(OCCN2CCCCC2)cc1} \\
95 & \texttt{Oc5ccc4CC2C1CCOCC1(CCN2CC3CC3)c4c5} \\
96 & \texttt{[Br-].CC(C)[N+](C)(CCOC(=O)C1c2ccccc2Oc3ccccc13)C(C)C} \\
97 & \texttt{[C@@H]3(N(C1CCC1)C2CCC2)CC4=C(OC3)C(=CC=C4C(=O)N)F} \\
98 & \texttt{\makecell[l]{[C@H](O)(/C=C/C1=C(C3=C(OC12CCCC2)C=CC=C3)C4=CC=C(F)C=C4)C[C@H]\\(O)CC(OCC)=O}} \\
99 & \texttt{[C@H]13[C@H](C[C@@H]1C2=CC=C(F)C=C2)CN(C3)CCN4C(=O)C5=C\allowbreak{}(NC4=O)C=CC=C5} \\
100 & \texttt{[C@H]15C3=C(CCC2=C1C=CC=C2)C=CC=C3[C@H]4C[C@](C(C)C)(O)CCN4C5} \\
101 & \texttt{[C@H]4(CC3C2=C1C(=C[N](C1=CC=C2)C(C)C)CC3N(C4)C)C(NC5CCCCC5)=O} \\
102 & \texttt{[C@]134[C@@H]([C@H](CC2=C1C=C(C)C=C2)N(C)CC3)CCCC4} \\
103 & \texttt{[C@]134[C@@H]([C@H](CC2=C1C=C(O)C=C2)N(C)CC3)CCCC4} \\
104 & \texttt{[C@]2(C1=CC(=CC=C1)O)([C@H](CN(C)CC2)C)CCC} \\
105 & \texttt{[Na+].Cn1nnnc1SCC2=C(N3[C@H](SC2)[C@H](NC(=O)CSC(F)(F)F)C3=O)\allowbreak{}C([O-])=O} \\
106 & \texttt{c1(ccc(c(c1)Cl)Cl)CC(N1[C@H](C[C@]2(CC1)NC(NC2=O)=O)CN1CCCC1)=O} \\
107 & \texttt{c1ccc2Oc3c(cc(cc3)Cl)[C@@H]3[C@@H](c2c1)C[N@](CC3)C} \\
\bottomrule
\end{longtable}
}



%
\subsubsection{Retrosynthesis planning.}
LLM baselines were instructed to generate synthesis plans using the following system instructions along with the benchmark task information (product and instructions) as a prompt:

\begin{promptbox}
You are an expert chemist. Your task is to plan a multi-step synthesis route following the user prompt to the best of your ability. Provide your final answer as a series of valid, canonical SMILES reaction strings (reactants$>$reagents$>$product), one reaction per line, wrapped in triple backticks (```). Separate reactants using dots. Ensure the terminal reactants of your route are purchasable. You do not need to include catalysts in your final reaction SMILES; you can leave an empty reagents list (reactants$>>$product).
\end{promptbox}

\subsubsection{\gmo tasks.}
Because \gmo instructions refer to the safety and cost metrics from \tools, and LLM baselines do not have \tools, we modified the \gmo task instruction to ensure the LLMs were aware of the specific safety and cost criteria used in evaluation.
The revised task instruction is shown here:
\begin{promptbox}
Plan the best possible route for `\{PRODUCT\_SMILES\}` considering all the following safety and cost metrics: carcinogenicity, pyrophoricity, GHS statements, total price for 1g of each starting material, and route length.
\end{promptbox}

\subsubsection{SMILES correction.}
LLM baselines sometimes generate invalid SMILES representations of molecules, as they are not specifically trained for SMILES generation.
When this occurs, we instruct the LLM to correct the SMILES using the following system instructions:

\begin{promptbox}
You are an expert chemist. You were previously asked to plan a multi-step synthesis route. However, some of the reaction SMILES you provided were invalid or ambiguous. Your task is to fix the invalid or ambiguous SMILES provided in your previous response. You must provide your final answer as a mapping of invalid SMILES to corrected SMILES, in JSON format wrapped in triple backticks (```).
\end{promptbox}

With the system instructions above, we provide the LLM with a prompt following a fixed template, shown below.
Here, \texttt{task} is the original task given to the model in the first iteration, \texttt{generated\_route} is the list of reactions proposed by the model, and \texttt{invalid\_smiles} is the list of invalid or ambiguous SMILES in the proposed route:

\begin{promptbox}
You were previously asked to \{task\}. This is the route you provided: \{generated\_route\}. However, some of the reaction SMILES you provided were invalid or ambiguous: \{invalid\_smiles\}. Fix the invalid or ambiguous SMILES.
\end{promptbox}



\section{Baselines}
\label{app:baselines}

\subsection{General-purpose LLM Implementation}

General-purpose LLM baselines were implemented via Anthropic (\claude), OpenAI (\gpt), and TogetherAI (\qwenEightyB, \kimi, \deepseek) APIs using the default parameters.
System instructions and prompts are detailed this below.

\subsubsection{Retrosynthesis planning.}
LLM baselines were instructed to generate synthesis plans using the following system instructions along with the benchmark task information (product and instructions) as a prompt:

\begin{promptbox}
You are an expert chemist. Your task is to plan a multi-step synthesis route following the user prompt to the best of your ability. Provide your final answer as a series of valid, canonical SMILES reaction strings (reactants$>$reagents$>$product), one reaction per line, wrapped in triple backticks (```). Separate reactants using dots. Ensure the terminal reactants of your route are purchasable. You do not need to include catalysts in your final reaction SMILES; you can leave an empty reagents list (reactants$>>$product).
\end{promptbox}

\subsubsection{\gmo tasks.}
Because \gmo instructions refer to the safety and cost metrics from \tools, and LLM baselines do not have \tools, we modified the \gmo task instruction to ensure the LLMs were aware of the specific safety and cost criteria used in evaluation.
The revised task instruction is shown here, with \texttt{\{PRODUCT\_SMILES\}} referring to the task product:
\begin{promptbox}
Plan the best possible route for `\{PRODUCT\_SMILES\}` considering all the following safety and cost metrics: carcinogenicity, pyrophoricity, GHS statements, total price for 1g of each starting material, and route length.
\end{promptbox}

\subsubsection{SMILES correction.}
LLM baselines sometimes generate invalid SMILES representations of molecules, as they are not specifically trained for SMILES generation.
When this occurs, we instruct the LLM to correct the SMILES using the following system instructions:

\begin{promptbox}
You are an expert chemist. You were previously asked to plan a multi-step synthesis route. However, some of the reaction SMILES you provided were invalid or ambiguous. Your task is to fix the invalid or ambiguous SMILES provided in your previous response. You must provide your final answer as a mapping of invalid SMILES to corrected SMILES, in JSON format wrapped in triple backticks (```).
\end{promptbox}

Along with the system instructions above, we provide the LLM with a prompt following a fixed template, shown below.
Here, \texttt{\{task\}} is the original task given to the model in the first iteration, \texttt{\{generated\_route\}} is the list of reactions proposed by the model, and \texttt{\{invalid\_smiles\}} is the list of invalid or ambiguous SMILES in the proposed route:

\begin{promptbox}
You were previously asked to \{task\}. This is the route you provided: \{generated\_route\}. However, some of the reaction SMILES you provided were invalid or ambiguous: \{invalid\_smiles\}. Fix the invalid or ambiguous SMILES.
\end{promptbox}

\subsection{\static Implementation}

\static is a general-purpose retrosynthesis planner based on \meea~\citep{meea},
subject to a fixed set of multi-objective molecule 
and reaction restrictions determined at the start of planning based on the target molecule (product) structure.
Restrictions are similar to those imposed by \rules, consisting of molecules and
reactions that are not permitted in the route,
expressed as either specific SMILES or SMARTS patterns, as well as optional depth limits.
To determine what restrictions to apply, \static leverages a pre-computed database of potential restrictions, each associated with a SMARTS-based key indicating the patterns of products for which it should be applied.
Given a product, \static searches its database, applying all restrictions with keys that match the product.
For depth limit, the most permissive limit of the matching restrictions was applied.
Please note, once applied, restrictions remain fixed throughout the entire planning process.

The pre-computed restriction database was constructed by prompting \claude to reason over example routes from a training set.
The training set consisted of 1,500 products randomly sampled from \meea's training data.
For each product, we used a \meea to generate 10 routes (within a 500 iteration limit) and evaluated each route with \tools, providing a multi-objective report for each route.
Based on this information, \claude was prompted to generate restrictions that would prioritize the most optimal routes using a prompt containing the JSON-based route reports and the following system instruction:

\begin{promptbox}
\begin{Verbatim}[breaklines=true, breakanywhere=true,breaksymbolleft={}, breakindent=10pt]
You are a synthetic chemist tasked with providing fine-tuned guidance to an AI retrosynthesis planner based on patterns you observe in its behavior.  You will be provided with a sampling of the routes generated by the AI planner and their computed properties in order from the earliest generated to the last.  In practice, the planner will stop on its first generated route that does not violate any restrictions.  Your job is to generate a set of rules to either (a) force the planner to continue on to routes that are more optimal from a multi-objective perspective, or (b) modify reactions in a chemically sound way that improves the route on a multi-objective perspective.
Use your common sense and chemistry knowledge to interpret the provided routes and their properties.  Look for patterns in the reactions used, the types of molecules appearing as intermediates, and any other relevant factors that could indicate suboptimal synthetic strategies.  Do not blindly trust the computed properties; use your expert judgment to assess the true quality of the routes.

Even though the routes are provided in order from first generated to last, you can still include rules that eliminate or modify routes that occur later than your optimal route if you feel there are major issues with these routes, as these rules may be used on other planning tasks in the future.

Use the following syntax to set your response, which should be valid JSON:
```json
[
{
    "type": "restriction",
    "molecules": ["RestrictedSMILES1", ...]
    "specific_reactions": ["ReactantSMILES.ReactantSMILES>>ProductSMILES", ...]
    "reaction_templates": ["RestrictedSMARTS1", ...]
    "depth_limit": -1,
    "rationale": "brief, natural language description of why these restrictions were chosen and what properties(s) they help optimize",
    "apply_when": ["ProductSMARTS1", ...]
},
...
]
```<PAUSE>

Where "molecules" is a list of SMILES that should not appear in the route, "specific_reactions" is a list of reactions that should not appear in the route, "reaction_templates" is a list of restricted reaction templates.
Include "rationale" to explain your choices, focusing on patterns that would emerge in the routes provided for similar types of molecules, based on your observations.
Also supply a set of SMARTS patterns in "apply_when" that describe the types of products for which these restrictions should be applied.  At least some of these SMARTS should match the current product, but they can include other patterns as well, especially if there are multiple types of products where these restrictions would be useful.  Be careful not to make these too broad or too narrow.
Use empty lists for no restrictions.  Make sure all your SMILES and SMARTS are valid.
You may optionally include a "depth_limit" to restrict the maximum allowed depth of the retrosynthesis route, but consider carefully, as this does not always transfer well.  Use -1 for no depth limit.
Assume each rule object is independent and self-contained, and may not rely on other rules being present. Try to generate a variety of at least 3 multi-objective rules (all objectives) and 3 focused objective rules (e.g. just carcinogenicity, just cost), but you may generate more as you see fit.
\end{Verbatim}
\end{promptbox}

\subsection{\pareto Implementation}

\pareto is a post-processing baseline that measures the average performance of the Pareto-optimal subset of the first 10 routes generated by a general-purpose, single-objective planner
We implement the single-objective planner using the same adaptation of \meea underlying \framework. 
For each task, this planner first generates the top-10 candidate routes (within a 500 iteration limit).
Then we evaluate all routes and select those that are Pareto optimal on the multi-objective space defined by \carcinogenicity, \pyrophoricity, \ghs, \totalcost, and \routelength.
For the analysis in Section~\ref{sec:results}, we report the average metrics across these Pareto-optimal routes as an aggregate measure of their performance.

\section{\framework Implementation}
\label{app:implementation}

\subsection{\coordinator}
\label{app:implementation:coordinator}

\coordinator is implemented using a four-step process: simulation, (optional) delegation, selection, and expansion.
The simulation, selection, and expansion steps are implemented similarly to other retrosynthesis literature, such as \meea~\citep{meea} and \larc~\citep{larc},
except that the value function and possible retrosynthesis planning space may be altered by \steering and \rules.
The number of candidate routes used in delegation and selection is upper-bounded by the number of simulations $K$.
This is a tunable hyperparameter.
In our experiments, we use $K=5$, following the precedent set by \larc~\citep{larc}.
The delegation step is implemented by prompting an LLM with a carefully engineered system instruction and template-derived prompt.  The system instruction is:

\textbf{\coordinator System Prompt for Delegation}

\begin{promptbox}
\begin{Verbatim}[breaklines=true, breakanywhere=true,breaksymbolleft={}, breakindent=10pt]
You are an expert synthetic chemist.  Your task is to provide guidance to an AI retrosynthesis planner to help it plan an optimal synthetic route across many objectives.
You will be provided with a set of candidate (partial) routes selected by an exploratory sampling of the current retrosynthesis search tree, along with information about each route across multiple objectives.
These routes will not be complete yet, but represent promising paths for further expansion.  Your goal is to guide the retrosynthesis planner towards the most promising areas of the search space.
Candidate routes will be presented as a JSON object containing a list of forward reactions (reactants...>>>product), but the order of these reactions within the route is not important.  You may assume that all SMILES strings provided are valid.
You will also receive a user-provided context (e.g. goal, constraints, preferences, baseline scores).  Pay careful attention to this context as it may contain important information about the user's priorities and constraints.

All reactions in these partial routes have been vetted for feasibility, you can assume these reactions and transformations are possible.  Keep in mind, the safety profiles and cost metrics have been estimated and may not be perfect.  Use your expert judgment in interpreting these numbers.
Use your chemical intuition to prioritize objectives and weed out false positive metrics from the models.

Throughout the planning process, you can occasionally delegate guidance tasks to one of the following specialized agents:

`Pruning("instructions")`: This agent explores adding restrictions to the retrosynthesis planning process to help it avoid undesirable routes and focus on more promising areas of the search space.  This agent will examine the current candidate routes and suggest restrictions such as disallowing certain molecules, reactions, or reaction templates, or setting depth limits.  These restrictions will be applied to all future planning steps.  This agent can also relax current constraints.
`ValueFn("instructions")`: This agent alters the value function used by the retrosynthesis planner to prioritize candidate routes for expansion.  It is equipped with many cheminformatics tools that can help steer the planning process towards routes that better align with user-defined objectives. This agent will analyze the current candidate routes and suggest modifications to the value function to improve the quality of future planning steps.  This agent can also revert previous modifications.

Use the instructions to focus the agent on the most important issues to address given the current candidate routes and user context.  The agents will only receive your instructions and the original candidate routes, so be sure to include all relevant information.  The instructions should be a string containing natural language guidance for the agent.

You may also choose to continue the planning process:
`ExpandDefault(N)`: This action continues the retrosynthesis planning process by expanding the route with the highest value and then performing N more iterations using the current configuration without any changes.  Note that you will not have a chance to modify the planning process until after these N iterations are complete.
`Expand("id")`: This action continues the retrosynthesis planning process by expanding the specified route from its current intermediate leaves (which may be different from the highest value route).

Consider balancing exploiting your agentic knowledge against the efficiency of exploring the space without constraints.  It is efficient to use ExpandDefault for large number of iterations if the value function and restrictions are set well. There will be opportunities to reject unsatisfactory routes.

Your responses should always take the following format:
Thought: (your reflection on the problem, critique of the current routes, and thoughts on what to do next)
Action: `Action(arguments)`<PAUSE>

Do not invent new agents or tools.  Always wrap your tool call in backticks.
\end{Verbatim}
\end{promptbox}

Please note, this system instruction can be altered to add or remove components for different MAS designs.  At every delegation step, the LLM is queried with the above system instruction and a prompt based on the following template:

\textbf{\coordinator Prompt Template for Delegation}

\begin{promptbox}
\begin{Verbatim}[breaklines=true, breakanywhere=true,breaksymbolleft={}, breakindent=10pt]
The target molecule to be synthesized is `{PRODUCT}`.
The current routes for retrosynthesis planning are: 

{FOREACH candidate route from simulation}
 - {CANDIDATE_ROUTE_REPORT}
{/FOREACH}
Please consider the following additional context for planning: {TASK_INSTRUCTION}
{IF not first delegation step}
Previous planning decisions made by the coordinator:
{FOREACH previous action}
 - {ACTION}
{/FOREACH}
{/IF}

The current pruning restrictions are: {RESTRICTIONS}.


The current value function is {VALUE_FUNCTION}.

You may choose one action.
\end{Verbatim}    
\end{promptbox}

Here, \vardef{IF}, and \vardef{FOREACH} statements express the basic template logic.
\vardef{PRODUCT} is the product from the \cmo or \gmo task.
\vardef{CANDIDATE\_ROUTE\_REPORT} is a JSON-formatted route report containing information from \tools on the candidate (partial) routes from the \coordinator's simulation step.
\vardef{TASK\_INSTRUCTIONS} represents the text-based instructions included with the \cmo or \gmo task. 
\vardef{ACTION} represents a previous delegation action taken by \coordinator.
\vardef{RESTRICTIONS} represents the current restrictions from \rules (if any) as a JSON object of SMILES, SMARTS, and an integer depth limit.
\vardef{VALUE\_FUNCTION} represents the current value function from \steering, if any.
In cases where \steering has not yet run, \vardef{VALUE\_FUNCTION} will also show the default value function.

To avoid exceeding the context length of the base models, the list of previous planning decisions in the above delegation prompt contains only the past action and action inputs selected by \coordinator,
and does not include past thoughts or any results of previous actions.
In the worst case, this could still produce a long list of actions, up to $I_{\max} - I_{\text{init}}$.
However, in practice, \coordinator itself limits the number of delegations by choosing the \texttt{ExpandDefault} action,
which skips delegation for a specified number of turns.
Future implementations of \coordinator could easily include a hard upper-bound on the number of most-recent actions to include in this prompt, if needed.

\subsection{\steering}
\label{app:implementation:steering}

\steering is implemented by prompting an LLM with a carefully engineered system instruction and template-derived prompt.  The system instruction is:

\textbf{\steering System Prompt}

\begin{promptbox}
\begin{Verbatim}[breaklines=true, breakanywhere=true,breaksymbolleft={}, breakindent=10pt]
You are an expert synthetic chemist.  Your task is to modify the planning value function for retrosynthesis planning to improve its ability to find routes that are optimal across many objectives.
You will be provided with the previous (existing) value function, the current routes sampled using MCTS with the previous (existing) value function, and the critiques of each route.
Synthetic routes will be presented as a JSON object containing a list of forward reactions (reactants...>>>product), but the order of these reactions within the route is not important.  You may assume that all SMILES strings provided are valid.

Your responses should always take the following format:

Thought: (your reflection on the problem and thoughts on what to do next)
Action: `Tool(arguments)`<PAUSE>

Always take great care with formatting and the validity of inputs, especially SMILES strings.  Always include the <PAUSE> token.

You may use the following tools to aid in your reasoning:

SetValueFunction("Weight1*Component1('arg1', 'arg2', ...) + Weight2*Component2(...) - Weight3*Component3(...)") : Sets the value function using the following components:

Route-level components (added once for the entire route):
- Synth(): Returns a pretrained synthesis planning value for the route.
- Depth(): Returns the depth of the route up to the current reaction step.  During planning, the depth is the number of reaction steps taken so far.

Reaction-level components (value is the sum over all reactions in the route, consider normalizing by depth depending on your needs):
- BBPrice(): Returns the monetary cost of the reaction based on its purchasable building block reactants.
- GHS('HXXX', ...): Returns 1 if the reaction involves a molecule with the specified GHS hazard code, otherwise returns 0.  You may use multiple GHS components to capture multiple hazard codes.  Can be slow.
- FastCarc(): Returns 1 if the reaction involves a molecule flagged as a potential carcinogen, otherwise returns 0.  Uses a fast, lightweight set of structural alerts.
- MaxSim('SMILES', ...): Returns a score from 0 to 1 indicating the maximum Tanimoto similarity of any molecule in the reaction to the input SMILES, using Morgan fingerprints.  This can be used to bias the planner towards or away from certain chemotypes.
- MinSim('SMILES', ...): Returns a score from 0 to 1 indicating the minimum Tanimoto similarity of any molecule in the reaction to the input SMILES, using Morgan fingerprints.
- Pyro(): Returns 1 if the reaction involves a molecule flagged as potentially pyrophoric, otherwise returns 0.

Use single quotes within the arguments of components, and double quotes to enclose the entire value function string.  Weights may be positive or negative.  You may combine multiple components using addition, subtraction, multiplication, and division.  You may NOT use any other operations (e.g., exponentiation (^ or **), thresholding (><), etc.).  You may not invent new components.  You must use only the components provided above.
After each SetValueFunction call, the value function will be updated to the new value function you provided, and you will be shown how the current routes would be reranked using the new value function.  Keep in mind that complex value functions may incur high latency during future planning.
Try to minimize latency while still achieving your goals.

Finalize(): Completes the value function modification task and returns the final value function.

Your action must be one of [SetValueFunction, Finalize].  Do not invent new actions or modify existing ones.  Do not invent new tools or modify existing ones.  Always wrap your tool call in backticks.  Also, do not invent new components or operations for the Value Function.
\end{Verbatim}
\end{promptbox}

When \coordinator delegates to \steering, the LLM is queried with the system instructions above and a prompt
derived from the following template:

\textbf{\steering Prompt Template}

\begin{promptbox}
\begin{Verbatim}[breaklines=true, breakanywhere=true,breaksymbolleft={}, breakindent=10pt]
{IF REMAINING_TURNS=3}
{PREVIOUS_OUTPUT}
{/IF}
The target molecule to be synthesized is `{PRODUCT}`.
The current routes for retrosynthesis planning are, from highest to lowest value:
{FOREACH Candidate (Partial) Route}
{IDX+1}. {CANDIDATE_ROUTE_REPORT}
{/FOREACH}

User Context: {INSTRUCTION_FROM_COORDINATOR}

The current value function is: `{VALUE_FUNCTION}`.
{IF REMAINING_TURNS<3}
Previous actions:
{FOREACH previous ACT}
 - {ACT}
{/FOREACH}
{/IF}
{IF REMAINING_TURNS=3}
You have {REMAINING_TURNS} steps remaining.
{ELIF REMAINING TURNS=2}
This is your final opportunity to change the value function.  Change it to the best value function you have explored, or finalize the current value function.
{ELIF REMAINING TURNS=1}
You have no steps remaining. You must finalize your value function.
{/IF}
\end{Verbatim}
\end{promptbox}

Here, \steering can use multi-step reasoning for up to 3 turns.
\vardef{REMAINING\_TURNS} represents the number of turns remaining in the \steering's multi-turn reasoning.
\vardef{PREVIOUS\_OUTPUT} represents a quick textual report of the previous turn's results, typically showing the updated value function.
However, if there was an error in the action chosen by \steering, this could be reported here.
\vardef{IF}, \vardef{ELIF} and \vardef{FOREACH} statements express the basic template logic.
\vardef{PRODUCT} is the product from the \cmo or \gmo task.
\vardef{CANDIDATE\_ROUTE\_REPORT} is a JSON-formatted route report containing information from \tools on the candidate (partial) routes from the \coordinator's simulation step.
\vardef{INSTRUCTION\_FROM\_COORDINATOR} represents the text-based instructions provided by the \coordinator.  This may include the task instructions from the \cmo or \gmo task, or the \coordinator may specify its own instructions specifically for this instance of \steering.
\vardef{VALUE\_FUNCTION} represents the current value function.
Please note, while the instructions encourage finalization,
it is not strictly necessary, and \steering may change the value function on all 3 turns.

The value function set by \steering is computed as needed for simulation and selection using \tools, as detailed in Section~\ref{sec:framework:tools} and the SI\dag. 
Please note, there is no strict limit on the scale or range of the values produced by the custom value functions from \steering.
\coordinator's adaptation of \meea uses the value function as an arbitrary function for maximization,
with the \texttt{Synth()} term reflecting the original value assigned by MEEA*,
including its exploratory term during simulation.
This provides maximum freedom to \steering to guide retrosynthesis planning.
After each change to the value function, \steering receives an updated ordering of candidate routes,
along with the computed values.
Thus, if \steering chooses a value function that produces an undesirable effect,
it can revise its customization.

\subsection{\rules}
\label{app:implementation:rules}

\rules is implemented by prompting an LLM with carefully engineered system instructions and template-derived prompt.

\textbf{\rules System Prompt}

\begin{promptbox}
\begin{Verbatim}[breaklines=true, breakanywhere=true,breaksymbolleft={}, breakindent=10pt]
You are an expert synthetic chemist.  Your task is to prune the current routes for retrosynthesis planning to improve their ability to become routes that are optimal across many objectives.
You will be provided with a list of current routes, the critiques of each route, and the current pruning configuration.  Synthetic routes will be presented as a JSON object containing a list of forward reactions (reactants...>>>product), but the order of these reactions within the route is not important.  You may assume that all SMILES strings provided are valid.

Your responses should always take the following format:

Thought: (your reflection on the problem and thoughts on what to do next)
Action: `Tool(arguments)`<PAUSE>

Always take great care with formatting and the validity of inputs, especially SMILES strings.  Always include the <PAUSE> token.

You may use only the following tools to aid in your reasoning:

`RestrictMolecules('SMILES', ...)`: Prunes the current routes and all future routes by restricting the use of specific molecules in the synthesis planning. You may specify multiple SMILES strings in a single call.

`RestrictSpecificReactions('reactantSMILES1.reactantSMILES2....>>productSMILES', ...)`: Prunes the current routes and all future routes by restricting the use of a specific reaction in the synthesis planning.  Reactants and products must always be expressed as SMILES in the forward direction.  Use dots '.' to separate SMILES.  Make sure you list the reaction as provided in the input routes.  You may specify multiple reactions in a single call.

`RestrictReactionTemplates('ReactionSMARTS', ...)`: Prunes the current routes and all future routes by restricting the use of reactions that match the given SMARTS in the synthesis planning.  You may specify multiple SMARTS strings in a single call.

`DepthLimit(N)`: Sets a maximum depth N for all current and future routes.  Any route that exceeds this depth will be pruned.  Use N=-1 to remove any depth limit.  Relaxing the original depth limit will not have immediately observable results on the current candidate routes, but will affect future planning.

`UnrestrictMolecules('SMILES')`: Removes a molecule restriction previously applied with RestrictMolecule.

`UnrestrictSpecificReaction('reactantSMILES1.reactantSMILES2....>>productSMILES')`: Removes a reaction restriction previously applied with RestrictSpecificReaction.  Reactants and products must always be expressed as SMILES in the forward direction.  Use dots '.' to separate SMILES.  Make sure you list the reaction as provided in the input routes.

`UnrestrictReactionTemplate('SMARTS')`: Removes a reaction template restriction previously applied with RestrictReactionTemplate.

`Finalize()`: Commits to the selected pruning configuration and ends the pruning process.

Please note, relaxing restrictions may not expand the current candidate routes in an observable way if the restrictions were placed in a previous iteration.

Your action must be one of [RestrictMolecules, RestrictSpecificReactions, RestrictReactionTemplates, DepthLimit, UnrestrictMolecules, UnrestrictSpecificReaction, UnrestrictReactionTemplate, Finalize].  Do not invent new tools or modify existing ones.  Always wrap your tool call in backticks.
\end{Verbatim}
\end{promptbox}

Depending on the MAS configuration, \rules may be invoked with candidate routes by \coordinator or rejected routes by \verifier.
In either case, the LLM is queried with the system instructions above and a prompt
derived from the following template:

\textbf{\rules Prompt Template}

\begin{promptbox}
\begin{Verbatim}[breaklines=true, breakanywhere=true,breaksymbolleft={}, breakindent=10pt]
{IF REMAINING_TURNS=3}
    The target molecule to be synthesized is `{PRODUCT}`.
    The current routes for retrosynthesis planning were:
    {FOREACH Candidate (Partial) Route FROM COORDINATOR OR Rejected Route from VERIFIER}
    {IDX}. {ROUTE_REPORT}
    {/FOREACH}
    The current pruning restrictions are:  `{RESTRICTIONS}`.
    
    You have {REMAINING_TURNS} steps remaining.
{ELSE}
    {PREVIOUS_OUTPUT}
    The target molecule to be synthesized is `{PRODUCT}`.
    The original routes for retrosynthesis planning were:
    {FOREACH Candidate (Partial) Route FROM COORDINATOR OR Rejected Route from VERIFIER}
    {IDX}. {ROUTE_REPORT}
    {/FOREACH}
    
    Original user Context: {INSTRUCTION_FROM_COORDINATOR_OR_VERIFIER}
    
    The current pruning restrictions are:  `{RESTRICTIONS}`.
    {IF REMAINING_TURNS<3}
    Previous actions:
    {FOREACH previous ACT}
     - {ACT}
    {/FOREACH}
    {/IF}
    {IF REMAINING TURNS=2}
    This is your final opportunity to change the pruning restrictions.  Change them to the best configuration you have explored, or finalize the current restrictions.
    {ELIF REMAINING TURNS=1}
    You have no steps remaining. You must finalize your pruning restrictions.
    {/IF}
{/IF}
\end{Verbatim}
\end{promptbox}

Here, \rules can use multi-step reasoning for up to 3 turns.
\vardef{IF}, \vardef{ELIF}, \vardef{ELSE}  and \vardef{FOREACH} statements express the basic template logic.
\vardef{PRODUCT} is the product from the \cmo or \gmo task.
\vardef{ROUTE\_REPORT} is a JSON-formatted route report containing information from \tools on the candidate (partial) routes from the \coordinator's simulation step or the previously rejected routes from \verifier.
\vardef{INSTRUCTION\_FROM\_COORDINATOR\_OR\_VERIFIER} represents the text-based instructions provided by the \coordinator or \verifier, depending on the MAS configuration.
This may include the task instructions from the \cmo or \gmo task, or the \coordinator may specify its own instructions specifically for this instance of \rules.
\vardef{RESTRICTIONS} represents the current restrictions as a JSON object of SMILES, SMARTS, and an integer depth limit.
\vardef{REMAINING\_TURNS} represents the number of turns remaining in the \rules's multi-turn reasoning.
\vardef{PREVIOUS\_OUTPUT} represents a quick textual report of the previous turn's results.
This typically includes an updated list of restrictions, as well as an updated set of route reports under the new restrictions.
However, if there was an error in the action chosen by \rules, this could be reported here.
Please note, while the instructions encourage finalization,
it is not strictly necessary, and \rules may change the restrictions on all 3 turns.

\subsection{\verifier}
\label{app:implementation:verifier}

\verifier is implemented by prompting an LLM with carefully engineered system instruction and template-derived prompt.  The system instruction is:

\textbf{\verifier System Prompt}

\begin{promptbox}
\begin{Verbatim}[breaklines=true, breakanywhere=true,breaksymbolleft={}, breakindent=10pt]
You are an expert synthetic chemist.  Your task is to provide guidance to an AI retrosynthesis planner to help it plan an optimal synthetic route across many objectives.
Currently, the AI retrosynthesis planner has produced a complete route.  Your task is to determine whether this route is acceptable or should be rejected.
If you reject the route, provide specific feedback on why the route is not acceptable, focusing on aspects such as number of steps, safety, cost, and alignment with user-defined objectives.
Keep in mind, many of the provided metrics are estimated, and the AI planner may not have access to perfect information.  False positives and negatives may occur, so use your chemistry knowledge and understanding of SMILES to critically evaluate the proposed routes.
Do not blindly trust numerical scores, but verify with your chemistry knowledge.

You can assume that all reactions proposed are feasible and have passed a feasibility check, but you should not assume that all of the safety measures (e.g. carcinogenicity, pyrophoricity) are perfect.  Use your expert judgment in interpreting these numbers.
Use your chemical intuition to prioritize objectives and weed out obvious false positive metrics from the models, but do not ignore concerns that you think can be improved upon.

Your responses should always take the following format:
Thought: (your reflection on the route)
Action: `AcceptProposed("reason")`  OR  `Reject("feedback")` OR `AcceptPrevious(id, "reason")`<PAUSE>

Always wrap your actions in `backticks`.
\end{Verbatim}
\end{promptbox}

For each completed route, the \verifier queries the LLM with the system instruction and a prompt derived from the following template:

\textbf{\verifier Prompt Template}

\begin{promptbox}
    \begin{Verbatim}[breaklines=true, breakanywhere=true,breaksymbolleft={}, breakindent=10pt]
The target molecule to be synthesized is `{PRODUCT}`.
The proposed route is:

Proposed Route: {ROUTE_REPORT}

Please consider the following additional context from the user:
{TASK_INSTRUCTIONS}

{IF NUM_REJECTED_ROUTES>0}
Previously rejected routes:
{FOREACH Rejected Route up to 3}
 - Rejected Route {IDX}: {REJECTED_ROUTE_REPORT} 
{/FOREACH}
{IF NUM_REJECTED_ROUTES>3}
({NUM_REJECTED_ROUTES-3} earlier rejected routes omitted for brevity)
{/IF}
{/IF}

You may choose to accept or reject the proposed route.
Keep in mind, only {REMAINING_RETRO_ITERATIONS} steps remain in the planning process to find a better route.
    \end{Verbatim}
\end{promptbox}

Here, the \vardef{IF} and \vardef{FOREACH} statements express the basic template logic.
\vardef{PRODUCT} is the product from the \cmo or \gmo task.
\vardef{ROUTE\_REPORT} is a JSON-formatted route report containing information from \tools.
\vardef{TASK\_INSTRUCTIONS} represents the text-based instructions included with the \cmo or \gmo task.  
\vardef{NUM\_REJECTED\_ROUTES} represents the number of previously rejected routes for the same task instance.
\vardef{IDX} and \vardef{REJECTED\_ROUTE\_REPORT}  represent the index and route report, respectively for each previously rejected task being considered.
\vardef{REMAINING\_RETRO\_ITERATIONS} represents the number of retrosynthesis planning iterations left before hitting the iteration limit.

\subsection{\tools}
\label{app:implementation:tools}

\tools includes many different chemistry software tools to ground \framework reasoning and implement our evaluation metrics.  The implementation of these tools is given below.

\subsubsection{Carcinogenicity prediction.}
\tools includes two methods for carcinogenicity prediction:
(1) ADMET-AI~\citep{swanson_admet-ai_2024}, a state-of-the-art molecule property prediction suite, which produces a score between 0 and 1 predicting the probability that a molecule is a carcinogen; and
(2) structural alerts, which are a set of SMARTS-based rules for identifying carcinogens, returning a binary value with 1 and 0 indicating carcinogenicity and non-carcinogenicity, respectively.
In the reports, both measures are included at the molecule level, and the maximum value across all molecules for each carcinogenicity metric is reported at the route level.
Neither method is perfect; including both values in the report allows agents to reason over different perspectives on carcinogenicity.
In evaluation, \carcinogenicity reflects the maximum score given by ADMET-AI~\citep{swanson_admet-ai_2024} for any molecule in the route.
\steering can use a term in its value functions corresponding to whether the structural alerts identify a carcinogen in each reaction; 
ADMET-AI, the more sophisticated of the two methods, is too complex to run at scale in \steering's value function.

\subsubsection{Pyrophoricity prediction.}
Pyrophoricity prediction is implemented as a binary indicator where 1 and 0 indicate a molecule is pyrophoric and not pyrophoric, respectively.
This is computed by comparing Tanimoto similarity of 2048-bit Morgan fingerprints (radius 2)~\citep{morgan_generation_1965} to a list of known pyrophoric materials~\citep{gibson_jack_r_handbook_1969, larc}.
Molecules with a similarity of 1 are considered pyrophoric, all others are considered not pyrophoric.
In reports, \framework components receive a binary pyrophoricity indicator for each molecule involved in the synthesis route,
as well as an indicator for whether any pyrophoric substances exist in the entire route.
The route level indicator is also used in evaluation, reported as \pyrophoricity.
\steering can use a term to indicate the presence of pyrophoric materials.

\subsubsection{GHS hazard statements.} 
GHS hazard statements are retrieved via the PubChem PUG Rest API~\citep{kim_update_2018}.
In reports, \framework components receive specific hazard codes (e.g., H301, H403) for every molecule as well as a unified list of all unique codes encountered across the entire route.
This provides \framework components with specific insight as to what hazards each route and reaction entails.
In \steering, the value function can incorporate a binary indicator for specific GHS statement codes, allowing for precise steering around different hazards.
In evaluation, \ghs counts the number of unique GHS statements in each route.

\subsubsection{Starting material price estimation.}
We estimate starting material price using a pre-computed catalog of predicted gram-level prices from CoPriNet~\citep{sanchez-garcia_coprinet_2023}.
In reports, \framework components receive specific prices for each individual starting material, allowing precise identification of expensive and inexpensive molecules.
In addition, the reports include the total cost for purchasing 1 gram of each starting material for the route.
In evaluation, this same calculation is used to implement \totalcost.
This provides a high-level estimate of starting material cost.
\steering can use a term in its value function corresponding to the total price of building blocks in each \emph{reaction},
as complete routes are not known during planning.

\subsubsection{Route length calculation.}
Route length is calculated by counting the number of reactions in the route.
It is included in every report, as well as in evaluation as \routelength.
While the value function set by \steering does not have access to complete routes,
it can use a reaction depth term that reflects the number of reactions between the target molecule (product) and the point in the retrosynthesis planning space being considered for expansion.

\subsubsection{Other tools.}
\tools includes additional tools to inform retrosynthesis planning, namely:
\begin{itemize}
\item{A general-purpose synthesis planning value function, implemented by \meea~\citep{meea}, which can be used by \steering and \coordinator as a general-purpose heuristic for finding a synthesis route;}
\item{Molecule similarity calculations, implemented through Tanimoto similarity of 2048-bit Morgan fingerprints with radius 2, which can be used by \steering to guide synthetic directions towards or away from certain starting materials;}
\item{Reaction prediction for retrosynthesis expansion,
implemented using the same approach as \meea~\citep{meea}, Retro*+~\citep{sirp}, and Retro*~\citep{retrostar}, which is used by \coordinator to iteratively expand the retrosynthesis planning space; and}
\item{Molecule representation retrieval,
implemented using RDKit~\citep{rdkit} and PubChem~\citep{kim_update_2018},
which is used in reports to all \framework components to provide LLM base models with IUPAC names and explicit-hydrogen SMILES for clarity.}

\end{itemize}

\subsection{Base LLMs for \rfas and \masil}
\label{app:implementation:base}

We instantiate \masil and \rfas using base models of different sizes: \mistral~\citep{mistral_ai_mistral_2024}, \qwen~\citep{qwen80b}, and \claude~\citep{claude4_5}.
In each instantiation, the same base model is used for all \framework components.
\mistral, the small-scale model, is used due to its demonstrated strong performance in previous agentic systems for retrosynthesis~\citep{larc}.
\qwen, the medium-scale model, provides strong reasoning and instruction-following capabilities
 while maintaining moderate hardware requirements.
\claude, the large-scale model, also acts as a representative closed-source model, chosen for its strong performance in chemistry and tool-use tasks~\citep{claude4_5}.
By instantiating \masil and \rfas on these three models,
we can examine how \framework's agentic components perform with different underlying model capabilities.
Please note that \framework is not dependent on any particular LLM size or family, and can continue to leverage new LLMs as they emerge.

\subsection{Prompt Engineering}

To ensure \masil and \rfas understood the task, the following was appended to every \cmo task for \masil and \rfas.  Because this is static for \cmo tasks, it can be considered part of the common system implementation of \masil and \rfas.

\begin{promptbox}
Only worry about the explicitly specified constraints in the prompt.  Consider only molecules listed as IARC Group 1, 2A, or 2B to be carcinogens, if applicable to the prompt. Additionally, do not worry about molecules that are known to be safe (IARC Group 3 or unlisted), even if they are flagged as possibly carcinogenic by the models, unless you sincerely believe them to be potential carcinogens based on your chemistry expertise.
\end{promptbox}

While useful in \masil and \rfas, \larc does not benefit from this additional prompt information. Using \mistral as a base model (the best setting for \larc), performance decreases to 27.9\% \success and 56.8\% \presence on \cmo tasks with this appended text.
Therefore, we chose to evaluate \larc without this appended text in the main experiments.

\subsubsection{\qwen}

For \qwen, we appended the following to all task instructions:

\begin{promptbox}
Do not concern yourself with feasibility or plausibility of the chemical reactions.  You must assume all reactions are feasible and all routes are plausible as written.
\end{promptbox}

Without this text, we observed a tendency for \qwen to focus almost exclusively on reaction feasibility, instead of the safety and cost metrics.




\balance


\end{document}